\documentclass{article}

\PassOptionsToPackage{numbers}{natbib}

\usepackage[main, final]{neurips_2026}
\usepackage{xcolor}
\usepackage{colortbl}
\usepackage{algorithm}
\usepackage{algorithmic}
\usepackage{amssymb}
\usepackage{enumitem}
\usepackage{tcolorbox}
\tcbuselibrary{listings,breakable,skins}
\usepackage{enumitem}

\usepackage{listings}
\usepackage{multirow}
\usepackage{booktabs}
\usepackage{wrapfig}

\usepackage[normalem]{ulem}
\useunder{\uline}{\ul}{}
\usepackage{subcaption}
\usepackage{caption}
\usepackage{bm}
\usepackage[british, american]{babel}
\newcommand{\tabcite}[1]{[\citeauthor{#1}, \citeyear{#1}]}

\definecolor{softred}{RGB}{200,120,120}
\definecolor{softgreen}{RGB}{120,170,120}

\usepackage{tikz}
\usetikzlibrary{shapes.geometric}

\definecolor{mygreen}{RGB}{0,150,0}
\definecolor{myred}{RGB}{200,0,0}

\definecolor{citecolor}{HTML}{0071BC}
\definecolor{linkcolor}{HTML}{ED1C24}

\usepackage[pagebackref=true,breaklinks=true,colorlinks,bookmarks=false,citecolor=citecolor,linkcolor=linkcolor,urlcolor=gray]{hyperref}

\usepackage[utf8]{inputenc}
\usepackage[T1]{fontenc}
\usepackage{url}
\usepackage{graphicx}
\usepackage{booktabs}
\usepackage{amsfonts}
\usepackage{nicefrac}
\usepackage{microtype}
\usepackage{amsmath}
\usepackage[capitalize]{cleveref}
\crefname{section}{Sec.}{Secs.}
\Crefname{section}{Sec.}{Secs.}
\crefname{subsection}{Sec.}{Secs.}
\Crefname{subsection}{Sec.}{Secs.}
\crefname{subsubsection}{Sec.}{Secs.}
\Crefname{subsubsection}{Sec.}{Secs.}
\crefname{figure}{Fig.}{Figs.}
\Crefname{figure}{Fig.}{Figs.}
\crefname{table}{Tab.}{Tabs.}
\Crefname{table}{Tab.}{Tabs.}
\crefname{equation}{Eq.}{Eqs.}
\Crefname{equation}{Eq.}{Eqs.}
\crefname{algorithm}{Alg.}{Algs.}
\Crefname{algorithm}{Alg.}{Algs.}
\usepackage{booktabs}

\usepackage{pifont}
\usepackage{placeins}  % \FloatBarrier: keep appendix floats within their section
\definecolor{ygreen}{RGB}{0,140,0}
\definecolor{yred}{RGB}{200,30,30}
\newcommand{\CT}[1]{\textcolor{ygreen}{{#1}}}
\newcommand{\DT}[1]{\textcolor{yred}{{#1}}}
\definecolor{linkpink}{RGB}{210, 80, 140}
\newcommand{\XMark}{\DT{Yes}}
\newcommand{\ARGen}{\textsuperscript{*}}

\newcommand{\eg}{\emph{e.g.}}
\newcommand{\ie}{\emph{i.e.}}
\newcommand{\vs}{\emph{vs.}\,}

\newcommand{\x}{\bm{x}}
\newcommand{\e}{\bm{\epsilon}}
\renewcommand{\v}{\bm{v}}

\newcommand{\z}{\bm{z}}
\newcommand{\s}{\bm{s}}
\newcommand{\cond}{\bm{c}}
\newcommand{\net}{\text{\texttt{net}}}
\newcommand{\mse}{\ensuremath{\mathcal{L}_{\textrm{MSE}}}}
\newcommand{\ce}{\ensuremath{\mathcal{L}_{\textrm{CE}}}}
\newcommand{\distill}
{\ensuremath{\mathcal{L}_{\textrm{distill}}}}
\newcommand{\reported}{\ensuremath{^{\dagger}}}
\newcommand{\reproduced}{\ensuremath{^{\ddagger}}}

\title{ELF: Embedded Language Flows}

\author{
\begin{tabular}{c}
\textbf{Keya Hu\textsuperscript{*} \quad Linlu Qiu\textsuperscript{*} \quad Yiyang Lu \quad Hanhong Zhao} \\[0.3em]
\textbf{Tianhong Li \quad Yoon Kim \quad Jacob Andreas \quad Kaiming He} \\[0.5 em]
\normalfont MIT \\
\normalfont \textsuperscript{*}Equal contribution; order decided by a coin flip. \\[0.3em]
{\hypersetup{urlcolor=linkpink}
Code: \url{https://github.com/lillian039/ELF}}
\end{tabular}
}

\begin{document}

\maketitle

\begin{abstract}
\vspace{-.5em}
Diffusion and flow-based models have become the \textit{de facto} approaches for generating continuous data, \eg, in domains such as images and videos. Their success has attracted growing interest in applying them to language modeling. Unlike their image-domain counterparts, today's leading diffusion language models (DLMs) primarily operate over discrete tokens. In this paper, we show that \textit{\mbox{continuous}} DLMs can be made effective with minimal adaptation to the discrete domain. We propose \textit{Embedded Language Flows (ELF)}, a class of diffusion models in continuous embedding space based on continuous-time Flow Matching. Unlike existing DLMs, ELF predominantly stays within the continuous embedding space until the final time step, where it maps to discrete tokens using a shared-weight network. This formulation makes it straightforward to adapt established techniques from image-domain diffusion models, \eg, classifier-free guidance (CFG). Experiments show that ELF substantially outperforms leading discrete and continuous DLMs, achieving better generation quality with fewer sampling steps. These results suggest that ELF offers a promising path toward effective continuous DLMs.

\end{abstract}

\begin{figure}[h]
\begin{minipage}[t]{0.48\textwidth}
\vspace{0pt}
\centering
\includegraphics[width=\linewidth]
{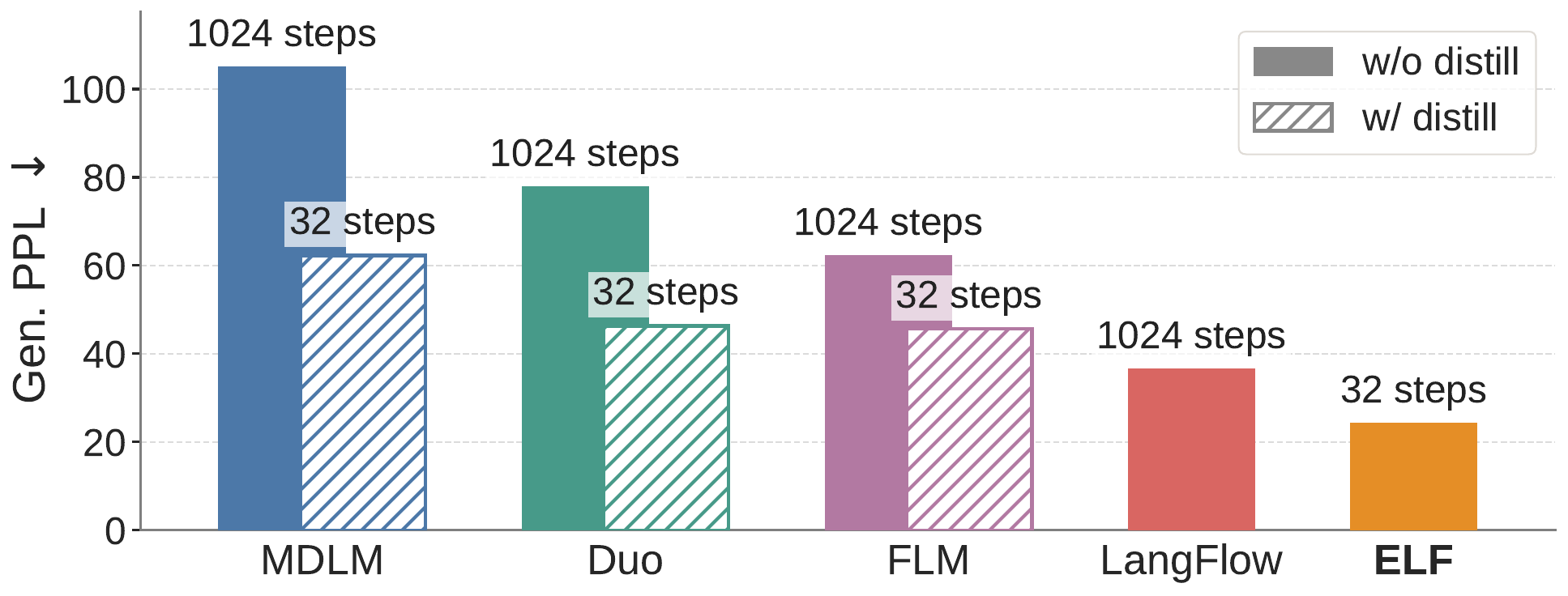}
\end{minipage}
\hfill
\begin{minipage}[t]{0.48\textwidth}
\vspace{0.5em}
\caption{\textbf{ELF} achieves lower generative perplexity with fewer sampling steps than prior DLMs, without using distillation. ELF achieves this while using $10\times$ fewer training tokens. (Model size: 105M for ELF and 170M for others; dataset: OWT. Detailed comparison in Fig.~\ref{fig:system_level_comparison}.)
}
\label{fig:sys}
\end{minipage}
\vspace{-1em}
\end{figure}

\section{Introduction}

Diffusion models \cite{sohl2015deep,song2020score,ho2020denoising} and flow-based models \cite{lipman2022flow,liu2022flow,albergo2023building} have become prominent paradigms for generating continuous data, demonstrating strong performance at synthesizing images, videos, and data in other continuous domains.
These advances have driven growing interest in extending diffusion methods to language modeling, leading to extensive work on diffusion language models (DLMs).
DLMs are commonly formulated in one of two ways: continuous or discrete.
Continuous DLMs map discrete tokens into continuous representations and perform denoising in the resulting continuous space~\citep{li2022diffusion,dieleman2022continuous,gong2022diffuseq}.
Discrete DLMs, in contrast, operate directly in token space and formulate a probabilistic diffusion model over discrete random variables~\citep{austin2021structured,he2023diffusionbert,lou2023discrete,sahoo2024simple,sahoo2025diffusion}.
Recent progress in DLMs has been mostly in the discrete regime, in large part due to the stronger empirical performance of discrete DLMs~\citep{li2025survey, nie2025llada, ye2025dream7bdiffusionlarge, sahoo2026scaling}.
But it remains an open question whether the current performance gap of continuous DLMs is due to the inherently discrete nature of language modeling or to underexplored algorithmic design choices.

In this work, we introduce Embedded Language Flows (\textbf{ELF}), a class of continuous DLMs based on Flow Matching~\citep{lipman2022flow,liu2022flow,albergo2023building}.
ELF is continuous in two senses.
First, it operates in \textit{continuous embedding} space by directly denoising continuous representations throughout the flowing process, with discretization considered only at the final time step. Second, it is formulated with \textit{continuous time}, following Flow Matching \citep{lipman2022flow,liu2022flow,albergo2023building}, which allows us to define the velocity field via the time derivative.
This formulation enables ELF to benefit from advances in Flow Matching, which is now widely used to instantiate diffusion models in image and video generation~\cite{ma2024sit,esser2024scaling,labs2025flux1kontextflowmatching,wan2025wan}.

\begin{figure}
\centering
\includegraphics[width=1.\linewidth]{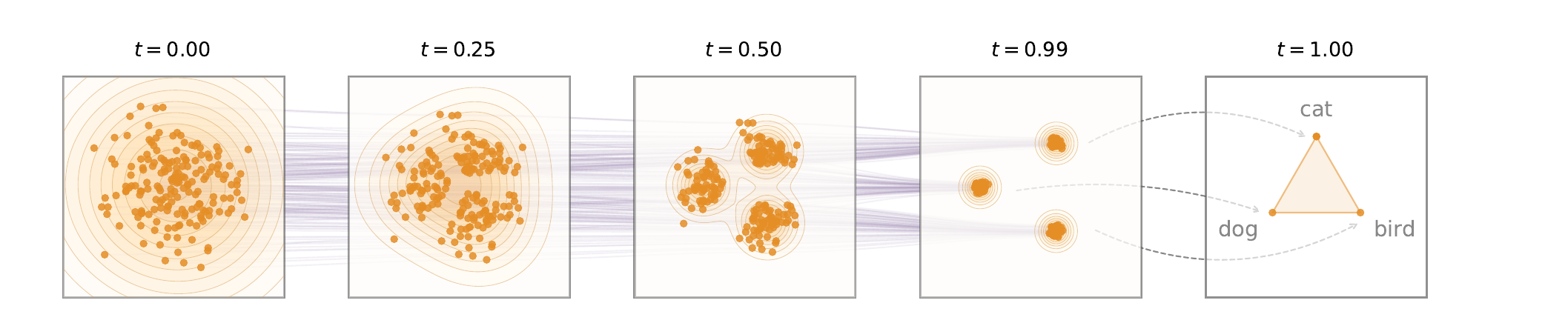}
\vspace{-1.5em}
\caption{
\textbf{Conceptual illustration of ELF.}
Orange points denote data represented in continuous embedding space, and purple lines show denoising trajectories from Gaussian noise to clean embeddings. Discretization is applied only at the final time step ($t=1$) using a shared-weight network.
}
\label{fig:elf_teaser}
\vspace{-1em}
\end{figure}

Following Latent Diffusion Models (LDM)~\cite{rombach2022high}, ELF constructs the continuous embedding space by applying an encoder model to the input discrete tokens.
The encoder can be pretrained, jointly trained, or frozen with random weights.
\textit{Unlike} latent diffusion, ELF does not require a separate decoder and thus introduces no additional component at inference time.
This design is based on the observation that the final time step in Flow Matching can be naturally repurposed to map continuous embeddings back to discrete tokens, eliminating the need for an explicit decoder.
As such, a shared-weight network is trained to perform denoising at all but the final step, and decoding (\ie, discretization) at the final step
(see Fig.~\ref{fig:elf_teaser}).

ELF builds on prior continuous DLMs, but aims for a minimalist design that addresses the interface between continuous and discrete spaces.
In contrast to pioneering work on continuous DLMs~\cite{li2022diffusion,dieleman2022continuous,gong2022diffuseq} and other approaches that employ a per-step discretization loss (\eg, cross-entropy), ELF performs denoising in continuous embedding space at nearly all steps, thereby offering maximal flexibility for the flow dynamics.
And unlike latent diffusion methods~\cite{lovelace2023latent,meshchaninov2025cosmos,shen2026codar}, which typically operate in a \textit{compressed} latent space and rely on a separate decoder, ELF directly operates in a high-dimensional latent space~\cite{li2025back} and requires no extra decoder.

Empirically, following established evaluation protocols, we show that ELF outperforms leading discrete DLMs and existing continuous DLMs (Fig.~\ref{fig:sys}).
ELF achieves better generation quality with fewer sampling steps than leading discrete DLMs (\eg, MDLM~\cite{sahoo2024simple} and Duo~\cite{sahoo2025diffusion}) and concurrent continuous DLMs (\eg, FLM~\cite{lee2026one} and LangFlow~\cite{chen2026langflow}).
Moreover, ELF achieves this performance using $10\times$ \textit{fewer} training tokens and \textit{without} any distillation.
We further show that ELF performs strongly on machine translation~\cite{wmt14} and summarization~\cite{Narayan2018DontGM}.
Overall, these results suggest that continuous DLMs can be highly competitive while requiring only minimal treatment of discretization, offering a promising direction for diffusion-based language modeling.

\section{Background \& Related Work}

\paragraph{Diffusion-/Flow-based models.} 
Diffusion models~\citep{sohl2015deep,ho2020denoising,song2020score} and flow-based models~\citep{lipman2022flow,liu2022flow,albergo2025stochastic} transform noise into data through ordinary or stochastic differential equations (ODEs/SDEs). In DDPM-style formulations, generation is defined by transitions between successive states~\citep{sohl2015deep,ho2020denoising,nichol2021improved}, which may be discrete or continuous. Discrete states require categorical transition distributions, as in discrete DLMs~\citep{austin2021structured,sahoo2024simple}; continuous states are commonly modeled through score or noise prediction under Gaussian corruption~\citep{song2020score,ho2020denoising,esser2024scaling}. Flow Matching extends this view to continuous time by learning the velocity field along a continuous path~\citep{lipman2022flow,liu2022flow,albergo2025stochastic}, where noise, data, and velocity predictions can be reparameterized into one another~\citep{esser2024scaling,li2025back}. Our method adopts Flow Matching to formulate language generation in continuous embedding space and continuous time.

\vspace{-0.5em}
\paragraph{Continuous diffusion language models.}
Continuous DLMs map discrete tokens to a continuous space to perform denoising. 
\textit{Embedding-space} methods, such as Diffusion-LM~\citep{li2022diffusion}, CDCD~\citep{dieleman2022continuous}, and DiffuSeq~\citep{gong2022diffuseq}, add Gaussian noise directly to token embeddings~\citep{strudel2022self, yuan2022seqdiffuseq, gulrajani2023likelihood, wang2023infodiffusion, ye2023dinoiser, lin2023text, wu2023ardiffusion, gao2024difformer}, with later work such as FlowSeq~\citep{hu2024flowseq} replacing the diffusion process with Flow Matching in the same embedding space.
A complementary direction studies \textit{simplex-based} representations, including SSD-LM~\citep{han2023ssd} and TESS~\citep{mahabadi2024tess,tae2025tess}, as well as related manifold-based formulations~\citep{jo2025continuous,davis2024fisher} and analog-bit encodings that denoise continuous binary codes of tokens~\citep{batzolis2026bitstream}.
Although these methods provide continuous relaxations of discrete tokens, their trajectories often remain tied to the discrete token space through mechanisms such as simplex constraints, token-aligned bit encodings, and token-level cross-entropy objectives.
In contrast, ELF denoises entirely in continuous embedding space without per-step token-level supervision and discretizes only at the final step.

Another line applies \textit{latent diffusion} to frozen encoder representations, represented by LD4LG~\citep{lovelace2023latent} and follow-up work~\citep{zhang2023planner,shabalin2025tencdm,lovelace2024dglm,meshchaninov2025cosmos,shen2026codar}, with recent work further learning the latent space jointly with the diffusion model~\citep{meshchaninov2026ldlm,guo2026cola}. Like many diffusion methods described above, these approaches typically follow DDPM-style or score-based formulations with DDPM noise schedules~\citep{ho2020denoising,nichol2021improved}, and additionally rely on a separate decoder network to recover tokens, trained either in an independent stage or jointly with the diffusion model~\citep{meshchaninov2026ldlm}. In contrast, ELF uses a continuous-time Flow Matching formulation with a linear (rectified-flow) interpolant~\citep{lipman2022flow,liu2022flow,albergo2025stochastic}, and does not require a separate decoder. This brings flow-based training and sampling into language diffusion, allowing ELF to benefit from recent advances in Flow Matching.

Several concurrent lines of work also revisit continuous flow-based language modeling. 
DFM~\citep{potaptchik2026discrete}, CFM~\citep{roos2026categorical}, FLM/FMLM~\citep{lee2026one}, and LangFlow~\citep{chen2026langflow} all incorporate token-level cross-entropy supervision along the flow trajectory, though they differ in the continuous state space, including simplex space, one-hot token encodings, and embedding space. 
Some of these methods further introduce distillation for few-step generation, such as distilled DFM/CFM and FMLM. 
In contrast, ELF keeps the denoising trajectory entirely in an unrestricted continuous embedding space, applying token-level supervision only at the final decoding step. 
A more comprehensive survey is provided in Appendix~\ref{sec:app_survey}.

\vspace{-0.5em}
\paragraph{Discrete diffusion language models.} 
Due to the discrete nature of language, another line of work applies diffusion directly in token space. 
D3PMs~\citep{austin2021structured} define general discrete corruption processes, including absorbing and uniform transitions. 
Masked diffusion models, such as MDLMs~\citep{sahoo2024simple}, use a special \texttt{[MASK]} absorbing state and generate samples through iterative unmasking~\citep{he2023diffusionbert,nie2025llada,ye2025dream7bdiffusionlarge,shi2024simplified}. 
Subsequent work improves sampling and efficiency through remasking, adaptive inference~\citep{wang2025remasking,wu2025fast}, and semi-autoregressive block diffusion, including E2D2~\citep{arriola2025encoder}.
Uniform-state diffusion models, such as Duo~\citep{sahoo2025diffusion}, instead diffuse tokens toward a uniform categorical distribution, enabling repeated token revision during inference~\citep{sahoo2025diffusion,deschenaux2026diffusiondualitychapterii,sahoo2026scaling}. 
Recent studies further scale discrete DLMs and extend them to code and multimodal generation~\citep{gong2025diffucoder,song2025seeddiffusionlargescalediffusion,yang2025mmada,you2025lladavlargelanguagediffusion,li2026omni}. A related line of work bridges discrete and continuous diffusion by jointly denoising in both the discrete and continuous spaces~\citep{gong2023diffuseqv2,zheng2025cadd,pynadath2025candi,zhou2026coevolutionary}. Overall, discrete diffusion models currently remain the dominant paradigm in diffusion-based language modeling~\citep{li2025survey}.
\section{Embedded Language Flows}
\begin{figure}
    \centering
    \includegraphics[width=0.90\linewidth]{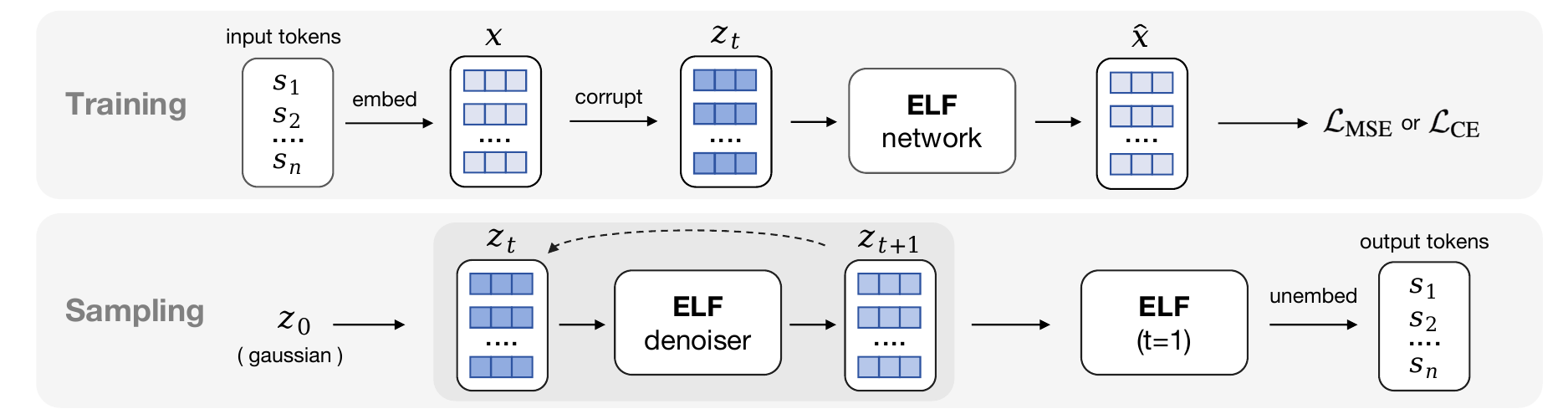}
    \caption{
    \textbf{During training}, discrete tokens are encoded into clean embeddings $\x$ and corrupted to $\z_t$, which ELF uses to predict $\hat{\x}$. The model is trained with either the denoising loss $\mathcal{L}_{\textrm{MSE}}$ or the token-wise cross-entropy loss $\mathcal{L}_{\textrm{CE}}$.
    \textbf{During inference}, ELF starts from Gaussian noise $\z_0$ and iteratively denoises embeddings from $\z_t$ to $\z_{t+1}$. Only at the final step does ELF switch to decoding mode and project the final embeddings back to discrete tokens through an unembedding layer.
    }
    \label{fig:method_details}
    \vspace{-1em}
\end{figure}

In this section, we present our flow-based formulation for language modeling (Fig.~\ref{fig:method_details}). Our method leverages the iterative nature of flow models to perform denoising primarily in continuous embedding space, converting clean embeddings back to discrete tokens only at the final step.
Following prior work~\citep{sahoo2024simple, sahoo2025diffusion, lee2026one, chen2026langflow}, we describe our method in the simpler setting of unconditional generation. The framework can be extended to conditional generation, as discussed in Sec.~\ref{sec:control}.

\subsection{The ELF Framework}
\paragraph{From discrete tokens to continuous embeddings.}
To apply continuous diffusion to language, we first map discrete tokens to continuous representations. Given a sentence, we tokenize it into a sequence of tokens $\s = [s_1, \ldots, s_L] \in V^L$, where each $s_i$ is drawn from the vocabulary $V$ and $L$ denotes the sequence length. We then map the discrete token sequence into a continuous embedding space. The choice of the embedding method is flexible. By default, we use a pretrained T5 encoder~\citep{raffel2020exploring} for bidirectional contextual embeddings. We also explore other jointly trained and randomized embeddings (see Sec.~\ref{sec:ablations}). The encoder is only used during training, which does not incur additional modules at inference. 

\vspace{-0.5em}
\paragraph{Flow Matching on continuous embeddings.}
\label{sec:flow_matching_sec}
After obtaining continuous language representations, we formulate the denoising process in the resulting embedding space using Flow Matching~\citep{lipman2022flow,liu2022flow,albergo2023building}.
Flow Matching defines a continuous flow path from noise to data in this space.
Let $\x \sim p_\text{data}(\x)$ denote the embedding distribution and $\e \sim p_\text{noise}(\e)$ denote the noise distribution (\eg, $\e\sim\mathcal{N}(0,\mathbf{I})$).
The noisy latent variable is defined by linear interpolation (``rectified flows''):
$\z_t = t \x + (1 - t)\e$,
where $t \in [0,1]$, and $\z_0 \sim p_\text{noise} $ and $\z_1 \sim p_\text{data}$.
In continuous time, the flow velocity $\v$ is defined as the time derivative of $\z$, that is, $\v = d\z/dt = \x - \e.$

While standard Flow Matching directly parameterizes $\v$ via a neural network, ELF follows recent advances on image generation and instead parameterizes $\x$ \cite{li2025back} (\textbf{$\x$-prediction}).
Specifically, let $\x_\theta = \net_\theta(\z_t, t)$ denote the network's immediate output. We train the model by minimizing the mean squared error (MSE) between the predicted velocity and the target velocity:
\begin{equation}
\mse = \mathbb{E}_{t,\x, \e} \| \v_\theta(\z_t, t) - \v \|^2
= \mathbb{E}_{t, \x, \e} \frac{1}{(1-t)^2}\| \x_\theta(\z_t, t) - \x \|^2,
\label{eq:loss}
\end{equation}
where we leverage the relation $\v(\z_t, t) = (\x - \z_t) / (1 - t)$ \cite{li2025back}.

The $\x$-prediction parameterization is important for ELF.
First, it enables Flow Matching to perform effectively on high-dimensional representations (\eg, 768-d per-token embeddings), consistent with observations in~\citep{li2025back} (see Appendix~\ref{sec:prediction_type} for ELF's ablations on prediction targets).
Second, predicting clean embeddings (\ie, $\x$) aligns naturally with the objective of predicting clean discrete tokens at the final step (discussed next), whereas the standard $\v$-prediction in Flow Matching does not.
Although $\v$ can be predicted by a network and transformed into $\x$, the weight sharing that ties the denoising (MSE loss) and decoding (cross-entropy loss) objectives is compromised. 
Empirically, we observe that $\v$-prediction works poorly when weights are shared with the final discretization step.

\vspace{-0.5em}
\paragraph{Back to discrete tokens.}
As the generation output consists of discrete tokens, we convert the clean embeddings back into tokens at the final time step (\ie, at $t=1$). 
By considering the final time step of ELF naturally as continuous-to-discrete decoding, our method does not require a separate decoder (or equivalently, it can be thought of as a decoder sharing weights with the denoiser). 

The network input at this time step should be $\z_t$ in the limit $t \to 1$. But because $\z_t \to \x$ as $t \to 1$, we introduce a token-level corruption process at this final step to create a nontrivial training input, denoted as $\tilde{\z}$ (detailed in Appendix~\ref{sec:app_train}).
The same network $\net_\theta$ maps $\tilde{\z}$ to a clean embedding ${\x}_\theta(\tilde{\z})$, which is subsequently projected by a learnable ``unembedding'' matrix $W$ to obtain logits.
We minimize a per-token cross-entropy (CE) loss against the ground-truth tokens $\s$:
\begin{equation}
\ce = \mathbb{E}_{\tilde{\z}}\left[\text{CrossEnt}(W{\x}_\theta(\tilde\z), \s)\right].
\label{eq:ce_loss}
\end{equation}
The network $\x_\theta$ shares weights with that in Eq.~(\ref{eq:loss}) and is conditioned on a binary ``mode'' token (denoise or decode) in addition to the time condition $t = 1$.
At inference time, we evaluate $W {\x}_\theta(\z_t)$ only at the final step $t = 1$, and apply $\operatorname{argmax}$ to obtain a discrete token at each position.

\subsection{Pseudocode}

\definecolor{codeblue}{rgb}{0.25,0.5,0.5}
\definecolor{codesign}{RGB}{0, 0, 255}
\definecolor{codefunc}{rgb}{0.85, 0.18, 0.50}
\definecolor{codenetwork}{RGB}{106, 90, 205}
\lstdefinelanguage{PythonFuncColor}{
  language=Python,
  alsoletter={_},
  keywordstyle=\color{blue}\bfseries,
  commentstyle=\color{codeblue},
  stringstyle=\color{orange},
  showstringspaces=false,
  literate=*
    {+}{{\color{codesign}+ }}{1}
    {-}{{\color{codesign}- }}{1}
    {*}{{\color{codesign}* }}{1}
    {/}{{\color{codesign}/ }}{1}
    {sample_t}{{\color{codefunc}sample\_t}}{1}
    {randn}{{\color{codefunc}randn}}{1}
    {randn_like}{{\color{codefunc}randn\_like}}{1}
    {jvp}{{\color{codefunc}jvp}}{1}
    {stopgrad}{{\color{codefunc}stopgrad}}{1}
    {metric}{{\color{codefunc}metric}}{1}
    {def}{{\color{codefunc}def }}{1}
    {return}{{\color{codefunc}return }}{1}
    {mse_loss}{{\color{codefunc}mse\_loss}}{1}
    {encode}{{\color{codefunc}encode}}{1}
    {ce_loss}{{\color{codefunc}ce\_loss}}{1}
    {argmax}{{\color{codefunc}argmax}}{1}
    {get_time_schedule}{{\color{codefunc}get\_time\_schedule}}{1}
    {unembed}{{\color{codefunc}unembed}}{1}
    {corrupt}{{\color{codefunc}corrupt}}{1}
    {random}{{\color{codefunc}random}}{1}
    {uniform}{{\color{codefunc}uniform}}{1}
    {samplt_t}{{\color{codefunc}samplt\_t}}{1}
    {net}{{\color{codenetwork}net}}{1}
}

\lstset{
  language=PythonFuncColor,
  backgroundcolor=\color{white},
  basicstyle=\fontsize{8pt}{8.9pt}\ttfamily\selectfont,
  columns=fullflexible,
  breaklines=true,
  captionpos=b,
}

\newcommand{\safeColorbox}[2]{%
  \begingroup
  \setlength{\fboxsep}{0pt}%
  \colorbox{#1}{\strut #2}%
  \endgroup
}

\begin{figure*}[t]\captionsetup{hypcap=false}% one anchor per algorithm (avoids duplicate PDF destinations)
\centering

\begin{minipage}[t]{0.48\linewidth}
\begin{algorithm}[H]
\caption{{ELF}: training.\\
{\scriptsize Two-branch computation is batched, adding no extra training cost.}
}
\label{alg:elf-training}
\begin{minipage}{0.98\linewidth}
\begin{lstlisting}[language=PythonFuncColor, escapechar=`]
# net(z, t, mode): ELF network
# s: a sequence of discrete tokens

x = encode(s)
if uniform(0, 1) < threshold:
    # denoising branch
    t = sample_t()
    e = randn_like(x)
    z = t * x + (1 - t) * e
    v = x - e
    x_pred = net(z, t, mode="denoise")
    v_pred = (x_pred - z) / (1 - t)
    loss = mse_loss(v_pred, v)
else:  
    # decoding branch (t = 1)
    z = corrupt(x)
    x_pred = net(z, t=1, mode="decode")
    s_pred = unembed(x_pred)
    loss = ce_loss(s_pred, s)
\end{lstlisting}
\end{minipage}
\end{algorithm}
\end{minipage}
\hfill
\begin{minipage}[t]{0.48\linewidth}
\begin{algorithm}[H]
\caption{ELF: inference.\\
{\scriptsize We show ODE for simplicity. SDE sampler is also applicable.}
}
\label{alg:dlm-inference}
\begin{minipage}{0.98\linewidth}
\begin{lstlisting}[language=PythonFuncColor, escapechar=`]
# shape: shape of embedded sequences
# ts: sampling time schedule, from 0 to 1

z = randn(shape)
for i in range(len(ts) - 1):
    t = ts[i]
    dt = ts[i + 1] - ts[i]
    x_pred = net(z, t, mode="denoise")
    
    # convert x prediction to velocity
    v = (x_pred - z) / (1 - t)
    z = z + dt * v

# final step
h = net(z, t=1, mode="decode")

# unembedding
token_logits = unembed(h)
tokens = argmax(token_logits)
\end{lstlisting}
\end{minipage}
\end{algorithm}
\end{minipage}
\vspace{-1em}
\end{figure*}

The core concepts of ELF are summarized in Alg.~\ref{alg:elf-training} and Alg.~\ref{alg:dlm-inference} (detailed in Appendix Fig.~\ref{fig:method_details2}).

\vspace{-0.5em}
\paragraph{Training.} 
As in standard Flow Matching, ELF employs a single network $\net_\theta$ to model all time steps, conditioned on $t$. This includes the final time step $t = 1$, which uses different pre-processing (corruption) and post-processing (loss computation). For clarity, we illustrate this distinction using an explicit ``\texttt{if}'' branch in Alg.~\ref{alg:elf-training}. In practice, samples from both branches are processed within a \textit{single} batch, and masking is used to selectively apply the appropriate corruption and unembedding operations as well as the corresponding loss terms. The network is further conditioned on a binary ``mode'' token that indicates whether the operation is ``denoise'' or ``decode''.

\vspace{-0.5em}
\paragraph{Inference.} 
During inference, ELF iteratively transforms noisy samples into clean embeddings. 
Starting from $\z_0 \sim \mathcal{N}(0,\mathbf{I})$, ELF solves the ODE: ${d\z_t}/{dt} = \v_\theta(\z_t, t)$, which is approximated with a numerical (\eg, Euler) solver.
At the final time step $t=1$, we apply the network under the ``decode'' mode and perform unembedding and discretization.

Besides the ODE formulation, our method also supports an SDE-inspired sampler. 
The underlying SDE associated with Flow Matching can be derived following~\cite{ma2024sit}, where the dynamics can be interpreted as injecting infinitesimal noise at each step.
In practice, we adopt a simpler approximation to emulate this behavior: we inject small noise at each step while correspondingly shifting the time variable $t$ toward the noise regime (detailed in Appendix, Alg.~\ref{alg:sampler}).
For brevity, we refer to the resulting SDE-inspired sampler as the ``SDE'' variant, while noting that it primarily captures the per-step stochastic behavior.
We experimentally compare the ODE formulation with this SDE variant.

\subsection{Conditioning and Guidance}
\label{sec:control}

Controlling model generation is an important aspect of generative modeling.
In image diffusion models, classifier-free guidance (CFG)~\citep{ho2021classifier} has been established as a highly effective technique for steering the generated output.\footnotemark{}
CFG also enables a trade-off between generation quality and diversity.
Because CFG was originally formulated for continuous quantities (\eg, score functions or velocity fields), it is naturally applicable to ELF.
This stands in contrast to discrete counterparts, where CFG remains largely unexplored and has been shown less effective~\citep{lee2026one,potaptchik2026discrete}.

In the absence of class labels, we employ \textit{self-conditioning}~\citep{chen2022analog} to construct the conditioning signals required for CFG.
Given that self-conditioning is already a standard component in DLMs~\citep{yuan2022seqdiffuseq,dieleman2022continuous,strudel2022self,lovelace2023latent,mahabadi2024tess,shabalin2025tencdm,shabalin2026gaussian}, incorporating CFG introduces only marginal computational overhead. In what follows, we first describe the self-conditioning used in ELF and then introduce CFG.

\footnotetext{CFG was historically introduced for \textit{class}-conditional generation. However, the notion of a condition can be generalized to other inputs, \eg, a text prompt. We use CFG in this broader sense, as our setting does not involve class labels.}

\vspace{-0.5em}
\paragraph{Self-conditioning.}
In a standard Flow Matching model (\ie, without self-conditioning), a forward pass at a given time step yields a single prediction. We denote this prediction by $\hat{\x}'$ in our case, indicating that it corresponds to a prediction of the clean embedding $\x$.
During training, self-conditioning~\citep{chen2022analog} performs a second forward pass, conditioned on $\hat{\x}'$, which serves as an intermediate prediction. The output of the second pass, denoted as $\hat{\x}$, can be written as $\hat{\x}=\net_\theta(\z_t \mid \hat{\x}', t)$. This is implemented by concatenating $[\z_t, \hat{\x}']$ as the network input \citep{chen2022analog}.
During training, the model is conditioned on $\hat{\x}'$ with probability 50\%, and uses a null condition $\mathbf{0}$ otherwise (see Appendix, Fig.~\ref{fig:method_details2} for details). 
During inference, the model conditions on the prediction from the previous time step, thus introducing no extra forward passes for inference.

The intermediate prediction $\hat{\x}'$ serves as a condition for the network. As such, it can be treated as the conditioning signal $\cond$ in the application of CFG, introduced next.

\vspace{-0.5em}
\paragraph{CFG with self-conditioning.} 
CFG \citep{ho2021classifier} combines the unconditional and conditional predictions through a linear extrapolation.
Formally, given a conditioning signal $\cond$, CFG in Flow Matching defines a velocity field as $\v_{\textrm{cfg}}(\z_t \mid \cond) = \omega \v(\z_t \mid \cond) + (1 - \omega)\v(\z_t \mid \varnothing)$, where $\varnothing$ denotes the unconditional counterpart and $\omega$ is the guidance scale.
As discussed, our conditioning signal $\cond$ is obtained from self-conditioning.
In its original form \citep{ho2021classifier}, CFG is applied at inference time, requiring two forward passes per step.

To avoid inference-time overhead, we adopt \textit{training-time} CFG techniques~\citep{chen2025visual,tang2025diffusion,geng2025mean, geng2025improved} previously developed for image generation.
These methods use a single network pass to model $\v_{\textrm{cfg}}$ instead of $\v$ (in our case, $\x_{\textrm{cfg}}$ instead of $\x$).
Because ELF is formulated similarly to its image-generation counterpart, adapting it to training-time CFG is straightforward, further illustrating the advantages of our continuous-based formulation. The implementation details, following the form in \citep{geng2025mean, geng2025improved}, are in Appendix (Alg.~\ref{alg:denoiser_cfg}, \ref{alg:decoder_cfg},~\&~\ref{alg:inference_cfg}).

\vspace{-0.5em}
\paragraph{Extension to conditional generation.}
Thus far, we have presented our method in the setting of unconditional generation, as in prior work~\citep{sahoo2024simple, sahoo2025diffusion, lee2026one, chen2026langflow}.
Our method can be naturally extended to conditional generation, in which outputs are conditioned on an input sequence (\eg, a prompt).
In this setting, we prepend the clean embeddings of the conditioning sequence to the model input and preserve them without corruption during both training and inference.
The model can then condition on them through self-attention.

CFG remains applicable in the conditional setting. The conditioning $\cond$ now consists of both the self-conditioning and the prefix clean embeddings; the unconditional counterpart is obtained by zeroing out $\cond$.
Analogous to text-to-image generation~\cite{esser2024scaling}, CFG is effective in controlling generation quality in our scenario, which can be viewed as ``text-to-text'' generation.

\section{Experiments}
\label{sec:exp}

\paragraph{Dataset and evaluation.}
For unconditional generation, we follow the experimental design used in past work~\citep{sahoo2024simple,sahoo2025diffusion,lee2026one,chen2026langflow}.
We train on the OpenWebText (OWT) dataset~\citep{gokaslan2019OpenWeb}, which has around 9B tokens, and pack sequences to length $L=1024$. For evaluation, we generate 1,000 samples and report generative perplexity (Gen.\ PPL), \ie, the perplexity of generated samples under a pretrained GPT-2 Large model~\citep{radford2019language}, together with average unigram entropy as a measure of sample diversity.\footnote{We do not use validation perplexity, since likelihood evaluation for flow-based models can require additional likelihood-specific training~\citep{ai2026joint}.} In Appendix~\ref{sec:additional_generation_quality}, we additionally evaluate our method using MAUVE~\citep{pillutla2021mauve}, which measures the distributional similarity between generated text and the original dataset, and LLM-as-a-judge, which uses a large language model (LLM) to assess the quality of each sample across different dimensions.

For conditional generation, we consider machine translation, summarization, and mathematical reasoning. For machine translation, we use the WMT14 German-to-English (De-En) dataset~\citep{wmt14} with sequence length $L=128$ (condition length 64, target length 64; 144M total target tokens), and evaluate using BLEU~\citep{papineni2002bleu}. For summarization, we use the XSum dataset~\citep{Narayan2018DontGM} with sequence length $L=1088$ (condition length 1024, target length 64; 6M total target tokens), and report ROUGE-1 (R1), ROUGE-2 (R2), and ROUGE-L (R-L)~\citep{lin2004rouge}. For mathematical reasoning, we train on 11.8M synthetic problems from TinyGSM~\citep{liu2023tinygsm} with sequence length $L=512$ (condition length 192, target length 320; 1.96B total target tokens), and evaluate on GSM8K~\citep{cobbe2021training} test set by executing one generated Python solution per problem and comparing its output with the ground-truth answer. We do not use sequence packing for conditional generation.

\vspace{-0.5em}
\paragraph{Model.} We use contextual embeddings from a frozen pretrained T5-small encoder~\citep{raffel2020exploring} (35M) with embedding dimension 512. We use a bottleneck design that linearly projects embeddings into a lower-dimensional space of size 128, and then projects them back to the hidden size of the model~\citep{li2025back}.
We consider three model sizes: ELF-B (105M), ELF-M (342M), and ELF-L (652M), and use ELF-B as the default for ablations. Detailed configurations are shown in Appendix Tab.~\ref{tab:elf_scaling_hparams}.

\vspace{-0.5em}
\paragraph{Training and inference.}
We train our model using the Muon optimizer~\citep{keller2024muon} with a learning rate of $0.002$ and a batch size of 512. The model is trained for 5 epochs on OWT (around 95K steps), 100 epochs on WMT14 and XSum (around 880K and 40K steps, respectively), and 11 epochs on TinyGSM (around 250K steps).
Depending on the selected model mode, the network is trained with either the MSE loss in Eq.~(\ref{eq:loss}) (80\%) or the CE loss in Eq.~(\ref{eq:ce_loss}) (20\%).
During inference, we use the ODE or SDE sampler to generate samples.

\subsection{Ablations}
\label{sec:ablations}

We begin by ablating several key design choices of our model on the simpler setting of unconditional generation on OWT, using the default ELF-B model and a 64-step ODE Euler sampler unless otherwise specified. More ablation studies are shown in Appendix~\ref{sec:additional_ablations}.

\vspace{-0.5em}
\paragraph{Classifier-free guidance (CFG).} 
\begin{wrapfigure}[12]{r}{0.42\textwidth}
\vspace{-1.4em}
\centering
\includegraphics[width=0.85\linewidth]
{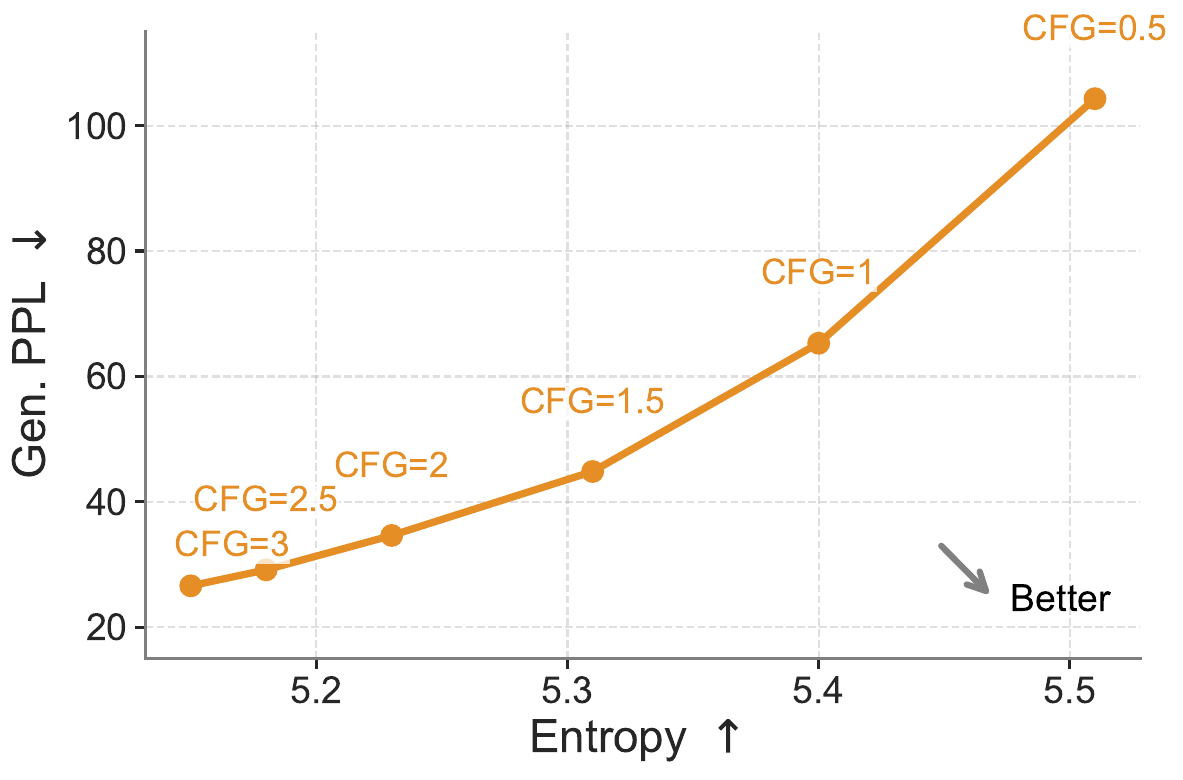}
    \vspace{-0.2em}
    \caption{\textbf{Ablations on guidance.}
    We evaluate the Gen. PPL--entropy trade-off across CFG scales: increasing the scale lowers Gen. PPL but reduces entropy.}
    \label{fig:cfg}
\end{wrapfigure}

Our flow-based continuous formulation is naturally compatible with CFG, a highly effective technique in standard diffusion models. Therefore, we first study the effect of the CFG scale. As shown in Fig.~\ref{fig:cfg}, increasing the CFG scale lowers generative perplexity but also reduces entropy, reflecting a quality--diversity trade-off. The preferred direction is toward the lower-right region of the plot, corresponding to lower generative perplexity and higher entropy. 
For most of the remaining ablations, we evaluate this quality--diversity trade-off by sweeping the CFG scale. Each point on the curve is computed from 1,000 generated samples at a specific CFG scale.

\vspace{1em}

\begin{figure}[h]
    \centering
    \includegraphics[width=\linewidth]{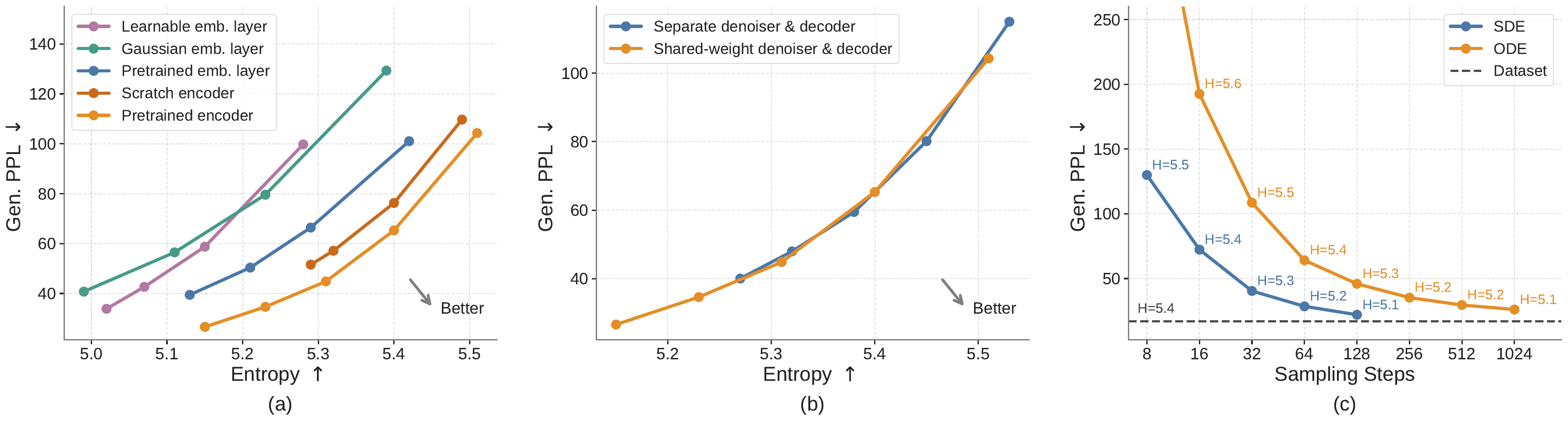}
    \caption{
    \textbf{Ablations on key design choices.}
    (a) Embedding choices: we compare contextual \vs non-contextual embeddings, as well as frozen \vs learnable embeddings; pretrained contextual embeddings achieve the best trade-off.
    (b) Decoding strategies: we compare a shared-weight denoiser-decoder with a two-stage, separately trained decoder. Both strategies achieve similar trade-offs, but the shared-weight variant extends further toward the regime of low generative perplexity.
    (c) Samplers: we compare ODE and SDE-inspired samplers across different sampling steps; SDE-inspired sampler consistently achieves lower generative perplexity in fewer steps.
    }
    \label{fig:ablations}
    % \vspace{-1em}
\end{figure}

\vspace{-0.5em}
\paragraph{Embedding choices.}
Since ELF operates in a continuous embedding space, we next study how the choice of embeddings affects performance. We ablate the continuous embeddings along two axes: whether the embeddings are contextual (\ie, from an encoder) or non-contextual (\ie, from a single embedding layer), and whether they are fixed or learnable. For contextual embeddings, we evaluate those from an off-the-shelf T5 encoder~\citep{raffel2020exploring} and embeddings from an encoder trained from scratch on OWT using the original T5 objective. For non-contextual embeddings, we consider token embeddings from the pretrained T5 model, frozen Gaussian embeddings, and learnable embeddings. See Appendix~\ref{sec:ablation_studies_details} for detailed setup.

We show the results in Fig.~\ref{fig:ablations}a. Contextual embeddings achieve a better generative perplexity--entropy trade-off. Embeddings from an encoder trained from scratch on OWT perform well, but slightly lag behind those from a pretrained encoder. Among the non-contextual variants, pretrained token embeddings outperform frozen Gaussian embeddings. Learnable embeddings perform the worst, likely due to the difficulty of jointly optimizing the embeddings and the denoiser. These results suggest that \textit{pretrained contextual embeddings} are favorable representations of language for ELF.

\vspace{-0.5em}
\paragraph{Decoding strategies.}
Since we use contextual embeddings as our continuous representations, we need to decode them back into discrete tokens. We use a shared-weight network, with training interleaving $\mse$ and $\ce$. Alternatively, we explore a two-stage strategy. In the first stage, we train a decoder from scratch with a frozen pretrained T5 encoder to reconstruct tokens from masked and noisy embeddings using $\ce$. In the second stage, we freeze both the encoder and decoder, and train a separate denoiser using $\mse$ (see Appendix~\ref{sec:ablation_studies_details} for details).
As shown in Fig.~\ref{fig:ablations}b, both strategies achieve a similar trade-off, but the shared-weight variant extends further toward the regime of low generative perplexity, while also simplifying the pipeline by avoiding an extra training stage.

\vspace{-0.5em}
\paragraph{Samplers.} Since ELF is formulated in continuous time and continuous space, it naturally supports both deterministic ODE sampling and stochastic SDE-like sampling; see Appendix Alg.~\ref{alg:sampler} for details. We compare ODE and SDE samplers across different sampling budgets with a self-conditioning CFG scale of 1. As shown in Fig.~\ref{fig:ablations}c, SDE sampling achieves substantially lower generative perplexity than ODE sampling in the few-step regime. 
These results suggest that introducing stochasticity during sampling can effectively reduce error accumulation and provide a better quality--efficiency trade-off.

\begin{wrapfigure}[15]{r}{0.4\textwidth}
    \vspace{-1.7em}
    \centering
    \includegraphics[width=0.9\linewidth]
    {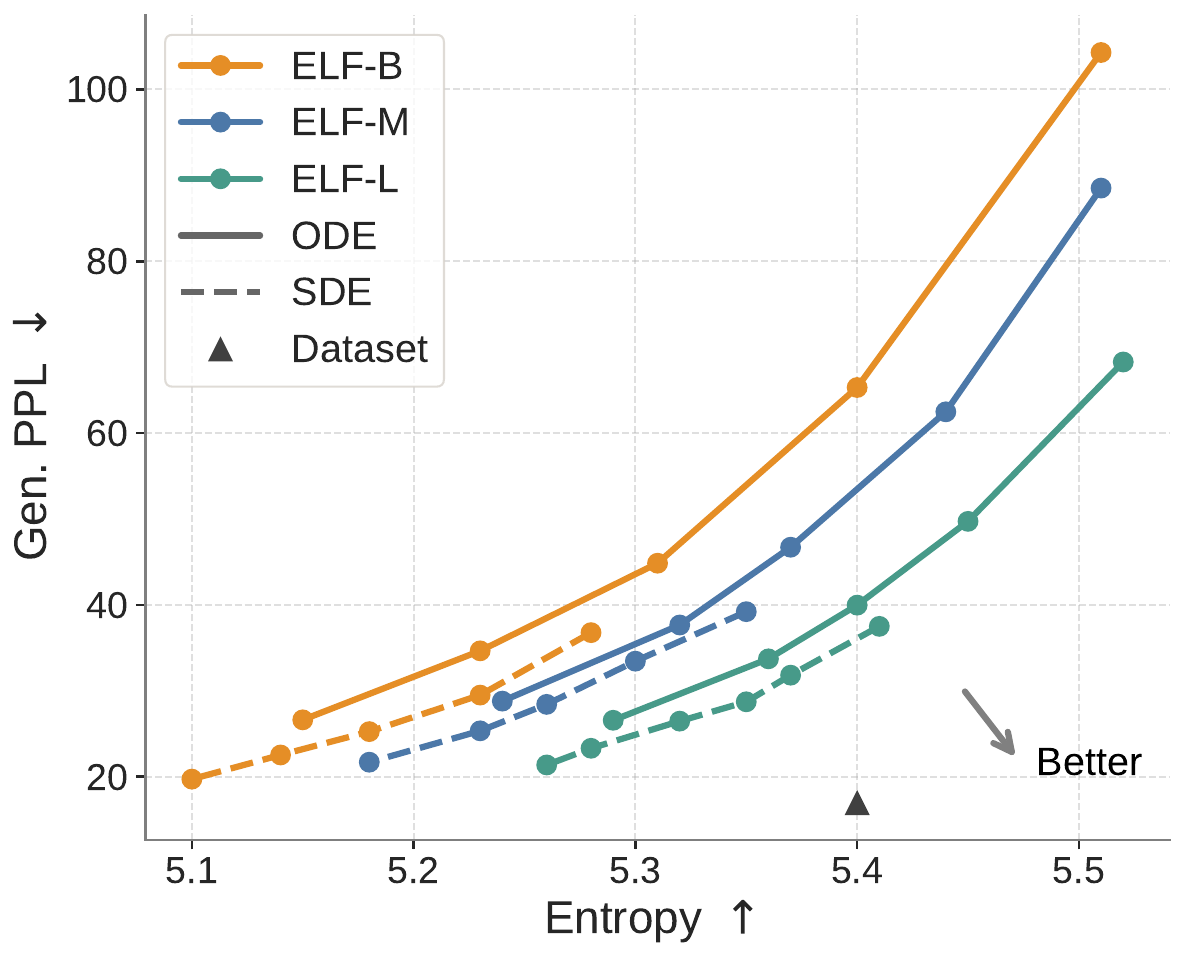}
    \caption{\textbf{Scaling of ELF models.}
    We compare ELF-B, ELF-M, and ELF-L. Scaling model size consistently improves the Gen. PPL--entropy frontier.}
    \label{fig:scaling}
    \vspace{-1em}
\end{wrapfigure}

% \vspace{-0.5em}
\paragraph{Model scales.}
We study the scaling behavior of ELF across three model sizes: \textbf{ELF-B} (105M), \textbf{ELF-M} (342M), and \textbf{ELF-L} (652M) (detailed in Appendix Tab.~\ref{tab:elf_scaling_hparams}). We evaluate each model using both ODE and SDE sampling. As shown in Fig.~\ref{fig:scaling}, scaling consistently improves the generative perplexity--entropy frontier. In particular, at matched entropy, larger models achieve lower generative perplexity, indicating higher sample quality with comparable diversity. 
Conversely, at similar generative perplexity, larger models maintain higher entropy.
The effect of the sampler is consistent across model sizes: SDE sampling improves over ODE sampling by pushing the frontier in a more optimal direction. These results suggest that ELF scales effectively, demonstrating the potential of model scaling. See Appendix Tab.~\ref{tab:elf_scaling_sampler_number} for the detailed numbers.

\vspace{1em}
\label{sec:system_uncond}
\begin{figure}[h]
    \centering
    \includegraphics[width=\linewidth]{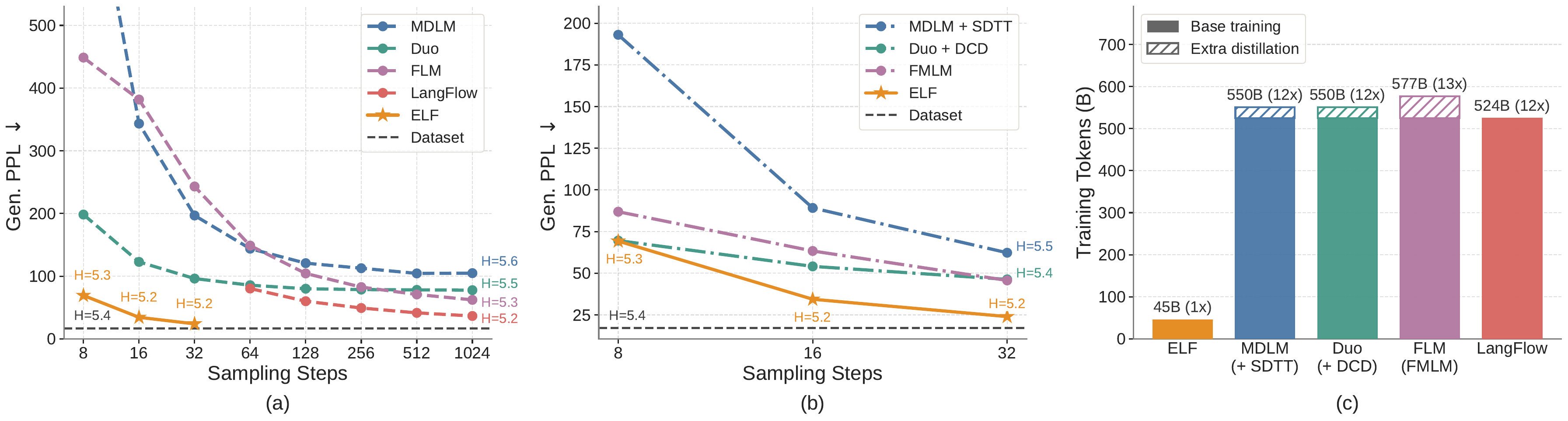}
    \vspace{-1em}
    \caption{
    \textbf{System-level comparison.}
    ELF-B outperforms both discrete and continuous DLMs trained under similar settings (a) and distilled variants of other baselines that require additional rounds of training (b), while using substantially fewer training tokens (c).
    }
    \label{fig:system_level_comparison}
\end{figure}

\subsection{System-Level Comparison on Unconditional Generation}
We first compare ELF-B against both discrete DLMs, including MDLM~\citep{sahoo2024simple} and Duo~\citep{sahoo2025diffusion}, and continuous DLMs, including FLM~\citep{lee2026one} and LangFlow~\citep{chen2026langflow}, under a comparable setting. All models are trained on the OWT dataset. ELF has 105M parameters, while the compared baselines have around 170M parameters. For ELF, we use our best configuration: SDE sampling with self-conditioning CFG scale of 3 (see Appendix~\ref{sec:app_hparams} for details). We show results in Fig.~\ref{fig:system_level_comparison}a. ELF achieves a generative perplexity of 24 using only 32 sampling steps, requiring substantially less inference-time compute than prior methods. 

ELF remains strong even compared with distilled models, which require extra training to distill a student model for few-step generation. As shown in Fig.~\ref{fig:system_level_comparison}b, in the few-step regime, ELF outperforms distilled models, including MDLM+SDTT~\citep{sahoo2024simple,deschenaux2024sdtt}, Duo+DCD~\citep{sahoo2025diffusion}, and FMLM~\citep{lee2026one}, even without any additional distillation. We further explore progressive distillation for ELF in Appendix~\ref{sec:distillation}. The distilled ELF model outperforms distilled baselines across 1--32 sampling steps.

ELF is also substantially more data-efficient in terms of estimated training tokens, as shown in Fig.~\ref{fig:system_level_comparison}c. While prior DLMs typically use over 500B tokens, ELF uses only 45B.\footnote{A per-method breakdown of training token counts is provided in Appendix Tab.~\ref{tab:training_tokens}. We also experimented with training on more tokens, but did not observe further performance improvement.} 
Together, these results show that, when combined with proper sampling and guidance, ELF achieves strong system-level performance. It not only improves inference efficiency, but also achieves strong performance with a much smaller training budget, demonstrating the potential of our flow-based language model. 
See Fig.~\ref{fig:qualitative_examples} for qualitative examples of ELF-B's generations. 

\subsection{System-Level Comparison on Conditional Generation}
\label{sec:cond}

\paragraph{Translation and summarization.}
\begin{table}[t]
\centering
\caption{
\textbf{Results on machine translation and summarization.}
We evaluate ELF-B on WMT14 German-to-English (De-En) translation and XSum summarization, comparing against baselines of similar parameter scale. \reported~denotes results taken directly from prior work and is the default source for De-En, while \reproduced~denotes results we reproduced using public codebases and is the default source for XSum. For XSum, we additionally report the standard error across evaluation examples when available. ELF achieves the best performance on both settings.
}

\label{tab:downstream}
\scalebox{0.8}{
\begin{tabular}{llcccc}
\toprule
\multirow{2}{*}{\textbf{Model}} & \multirow{2}{*}{\textbf{Size}} & \textbf{De-En}\reported & \multicolumn{3}{c}{\textbf{XSum}\reproduced} \\
 & & BLEU $\uparrow$ & ROUGE-1 $\uparrow$ & ROUGE-2 $\uparrow$ & ROUGE-L $\uparrow$ \\
\midrule
AR & 99M & {25.2} & 30.5 {\scriptsize$\pm$ 0.13} & 10.2 {\scriptsize$\pm$ 0.11} & 24.4 {\scriptsize$\pm$ 0.12} \\
MDLM~\cite{sahoo2024simple} & 99M & 18.4 & {33.4} {\scriptsize$\pm$ 0.11} & {11.6} {\scriptsize$\pm$ 0.10} & {25.8} {\scriptsize$\pm$ 0.10} \\
Duo~\cite{sahoo2025diffusion} & 170M (+35M) & 21.3\reproduced & 31.4 {\scriptsize$\pm$ 0.12} & 10.1 {\scriptsize$\pm$ 0.10} & 25.0 {\scriptsize$\pm$ 0.12} \\
E2D2~\cite{arriola2025encoder} & 99M & 24.8 & 28.4 {\scriptsize$\pm$ 0.11} & 8.3 {\scriptsize$\pm$ 0.09} & 22.0 {\scriptsize$\pm$ 0.10} \\
DiffuSeq~\cite{gong2022diffuseq} & - & - & 19.3\reported & 1.7\reported & 14.1\reported \\
CDCD~\cite{dieleman2022continuous} & - & 24.9 & - & - & - \\
\midrule
Ours & 105M (+35M) & \textbf{26.4} & \textbf{36.0} {\scriptsize$\pm$ 0.13} & \textbf{12.2} {\scriptsize$\pm$ 0.11} & \textbf{27.8} {\scriptsize$\pm$ 0.12} \\
\bottomrule
\end{tabular}
}
\vspace{-1.8em}

\end{table}

We compare ELF-B with autoregressive and diffusion-based baselines at a similar model scale. These include discrete DLMs (MDLM~\citep{sahoo2024simple}, Duo~\citep{sahoo2025diffusion}, and E2D2~\citep{arriola2025encoder}) and continuous DLMs (SeqDiffuSeq~\citep{yuan2022seqdiffuseq} and CDCD~\citep{dieleman2022continuous}). Some results are taken from the literature and others are reproduced from public codebases. See Appendix Tab.~\ref{tab:all_configs_full} for a summary.
We use the best sampling configuration selected on the validation set: a 64-step ODE sampler with the self-conditioning CFG scale set to 1 and the input-condition CFG scale set to 2.

\begin{figure}[t]
    \centering
    \includegraphics[width=1.0\linewidth]{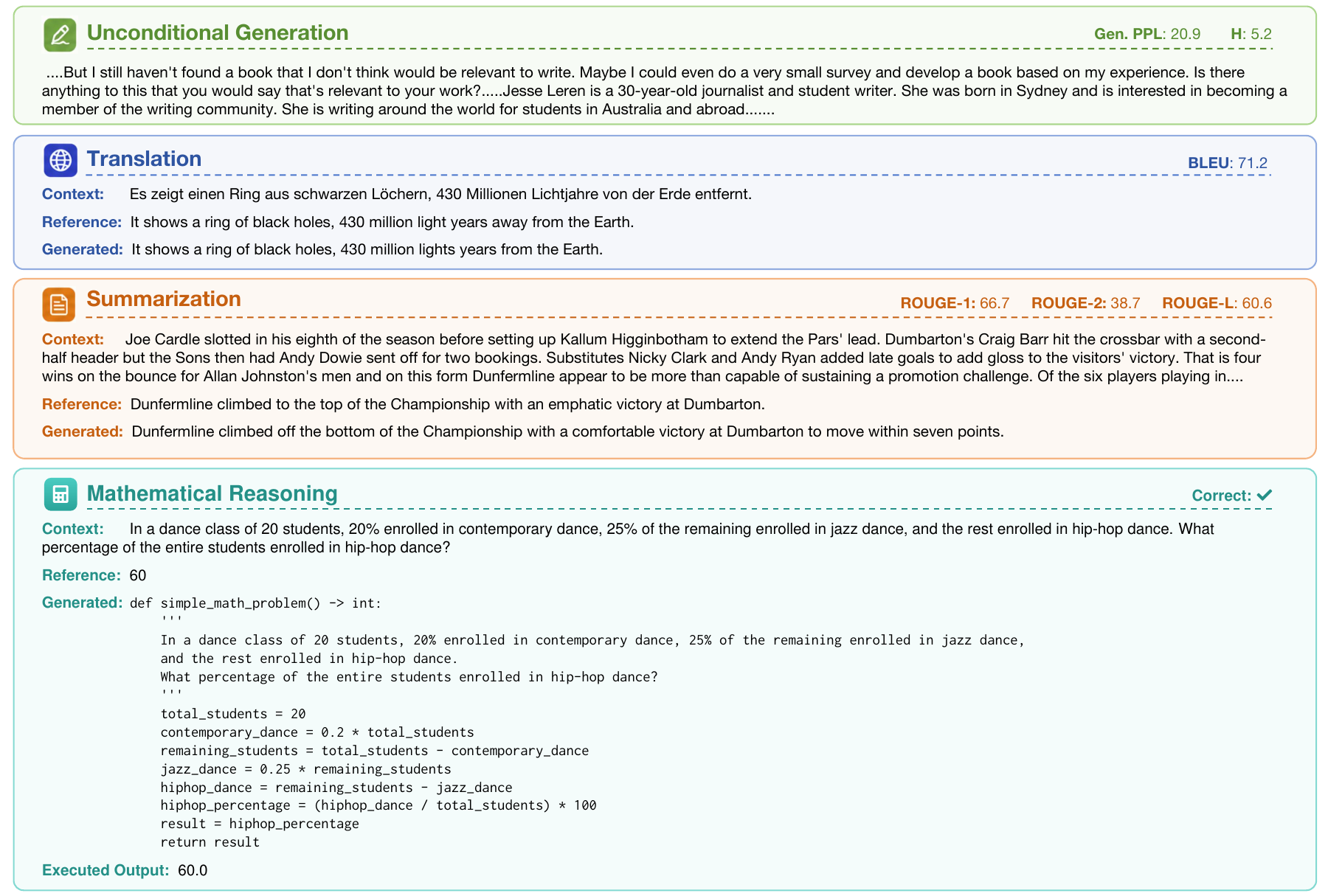}
    \vspace{-1em}
    \caption{\textbf{Qualitative examples} of text generated by ELF-B. We show an unconditional sample on OpenWebText, a German-to-English translation example, a summarization example, and a GSM8K mathematical reasoning example (the generated Python program is executed to obtain the answer). We also show their automatic evaluation metrics.
    Some text is omitted due to space limits; see Appendix~\ref{sec:qualitative_examples} for more examples.
    }
    \label{fig:qualitative_examples}
    % \vspace{-1.2em}
\end{figure}

We show the results in Tab.~\ref{tab:downstream}. ELF-B achieves the best performance among all compared methods on both tasks, demonstrating the effectiveness of ELF for conditional generation. Qualitative examples in Fig.~\ref{fig:qualitative_examples} further show that ELF-B generally follows the input context and generates outputs that are semantically aligned with the ground-truth references.

\vspace{-0.5em}
\paragraph{Mathematical reasoning.}
\begin{wraptable}[12]{r}{0.4\textwidth}
\vspace{-1.1em}
\centering
\caption{\textbf{Results on GSM8K accuracy (\%).} ELF substantially outperforms prior continuous and discrete DLMs with both sampling methods.}
\label{tab:gsm8k}
\scalebox{0.8}{
\begin{tabular}{lcc}
\toprule
\textbf{Model} & \textbf{ODE} & \textbf{SDE} \\
\midrule
MDLM~\citep{sahoo2024simple} & 18.0 & 33.0 \\
Duo~\citep{sahoo2025diffusion} & 17.2 & 36.0 \\
CANDI~\citep{pynadath2025candi} & 0.2 & -- \\
FLM~\citep{lee2026one} & 0.3 & -- \\
$\mathbb{S}$-FLM + top-$k=1$~\citep{deschenaux2026language} & 18.0 & 18.0 \\
\midrule
\textbf{ELF-B} & \textbf{39.7} & \textbf{40.8} \\
\bottomrule
\end{tabular}
}
\vspace{-1em}
\end{wraptable}

We evaluate ELF-B on GSM8K, a mathematical reasoning benchmark consisting of grade-school math problems, following~\citep{deschenaux2026language}. All models use 1,024 sampling steps. We use a CFG scale of 2.5 for both ODE and SDE sampling, and a noise re-injection scale of 0.2 for SDE. The baselines are evaluated without temperature scaling (the closest comparison to our ODE setting) and with low-temperature decoding (the closest comparison to our SDE setting). As shown in Tab.~\ref{tab:gsm8k}, ELF substantially outperforms prior continuous and discrete DLMs in both settings, demonstrating its potential for more practical and complex reasoning tasks.

\section{Conclusion}
We introduced \textbf{Embedded Language Flows} (ELF), a continuous diffusion language model that formulates language generation in continuous embedding space using continuous-time Flow Matching. In contrast to prior DLMs, ELF keeps the denoising trajectory continuous and applies discretization only at the final step, enabling straightforward adaptation of techniques from continuous diffusion models. Empirically, compared with leading discrete DLMs and existing continuous DLMs, ELF achieves a strong quality--efficiency trade-off across language generation tasks, attaining lower generative perplexity with fewer sampling steps and fewer training tokens. These results suggest that continuous DLMs remain a promising direction for diffusion-based language modeling.

\begin{ack}
We thank Mingyang Deng, Zhengyang Geng, Belinda Li, Itamar Pres, and Laura Ruis for their helpful feedback and insightful discussions.
We thank Google TPU Research
Cloud (TRC) for granting us access to TPUs. This work was supported in part by the Siegel Family Foundation Quest for Intelligence and MIT-IBM Computing Research Lab. The authors declare no competing interests.
\end{ack}

\newpage
\bibliography{elf}
\bibliographystyle{plainnat}
\newpage

%%%%%%%%%%%%%%%%%%%%%%%%%%%%%%%%%%%%%%%%%%%%%%%%%%%%%%%%%%%%

\appendix
\section{Continuous Diffusion Language Model Survey}
\label{sec:app_survey}

\begin{table*}[t]
\centering
\caption{
\textbf{Survey of continuous diffusion and flow-based language models.}
We summarize representative continuous diffusion and flow-based language models along several design axes.
\textit{Process} denotes the diffusion or flow process, with \CT{green} indicating continuous-time formulations and \DT{red} indicating discrete-time formulations.
\textit{State} denotes the continuous state in which denoising is performed.
\textit{Train per-step discr.} marks methods that convert intermediate denoising states to token predictions during training and apply token-level supervision such as cross-entropy loss at intermediate steps.
\textit{Infer.\ per-step discr.} marks methods that project intermediate sampling states back to token-aligned states during generation.
\textit{Sep.\ dec.} marks methods that require a separate decoder to map latent representations back to text.
Blank entries indicate absence.
\ARGen{} denotes autoregressive or block-autoregressive generation.
}
\label{tab:dlm-survey}
\scriptsize
\setlength{\tabcolsep}{4.2pt}
\renewcommand{\arraystretch}{1.02}

\begin{tabular}{l l l c c c}
\toprule
\multirow{3}{*}{\textbf{Method}}
& \multirow{3}{*}{\textbf{Process}\ensuremath{^{\dagger}}}
& \multirow{3}{*}{\textbf{State}\ensuremath{^{\ddagger}}}
& \textbf{Train}
& \textbf{Infer.}
& \multirow{3}{*}{\textbf{Sep. dec.}} \\
& &
& \textbf{per-step}
& \textbf{per-step}
& \\
& &
& \textbf{discr.}
& \textbf{discr.}
& \\
\midrule
\multicolumn{6}{l}{\emph{Embedding-space Diffusion LMs}} \\
\midrule
Diffusion-LM~\tabcite{li2022diffusion}         & \DT{DDPM}        & learn emb & \XMark & \XMark &        \\
SED~\tabcite{strudel2022self}                  & \DT{DDPM}        & fix emb   & \XMark &        &        \\
CDCD~\tabcite{dieleman2022continuous}          & \CT{Score-ODE}   & learn emb & \XMark &        &        \\
DiffuSeq~\tabcite{gong2022diffuseq}            & \DT{DDPM}        & learn emb & \XMark & \XMark &        \\
GENIE~\tabcite{lin2023text}                    & \DT{DDPM}        & learn emb & \XMark &  &        \\
AR-Diffusion~\tabcite{wu2023ardiffusion}\ARGen & \DT{DDPM}        & learn emb & \XMark &        &        \\
Plaid~\tabcite{gulrajani2023likelihood}        & \CT{VLB}         & learn emb & \XMark &        &        \\
InfoDiffusion~\tabcite{wang2023infodiffusion}  & \DT{DDPM}        & learn emb & \XMark &        &        \\
Difformer~\tabcite{gao2024difformer}           & \DT{DDPM}        & learn emb & \XMark &        &        \\
SeqDiffuSeq~\tabcite{yuan2022seqdiffuseq}      & \DT{DDPM}        & learn emb & \XMark &        &        \\
DINOISER~\tabcite{ye2023dinoiser}              & \CT{SDE/DDIM}    & learn emb & \XMark &        &        \\
FlowSeq~\tabcite{hu2024flowseq}                & \CT{FM}          & learn emb & \XMark &        &        \\
\midrule
\multicolumn{6}{l}{\emph{Simplex Diffusion LMs}} \\
\midrule
SSD-LM~\tabcite{han2023ssd}\ARGen              & \DT{DDPM}        & simplex   & \XMark & \XMark &        \\
TESS~\tabcite{mahabadi2024tess}                & \DT{DDPM}        & simplex   & \XMark & \XMark &        \\
RDLM~\tabcite{jo2025continuous}                & \CT{RDM}         & simplex   & \XMark &        &        \\
TESS\,2~\tabcite{tae2025tess}                  & \DT{DDPM}        & simplex   & \XMark & \XMark &        \\
Fisher-Flow~\tabcite{davis2024fisher}          & \CT{FM}          & simplex   &        &        &        \\
\midrule
\multicolumn{6}{l}{\emph{Latent Diffusion LMs}} \\
\midrule
LD4LG~\tabcite{lovelace2023latent}\ARGen       & \DT{DDPM}        & fix enc   &        &        & \XMark \\
PLANNER~\tabcite{zhang2023planner}\ARGen       & \CT{DDPM}        & fix enc   &        &        & \XMark \\
DGLM~\tabcite{lovelace2024dglm}\ARGen          & \CT{VP-DDPM}    & fix enc   &        &        & \XMark \\
TEncDM~\tabcite{shabalin2025tencdm}            & \CT{VP-DDPM}     & fix enc   &        &        & \XMark \\
Cosmos~\tabcite{meshchaninov2025cosmos}        & \CT{VP-DDPM}     & fix enc   &        &        & \XMark \\
CoDAR~\tabcite{shen2026codar}\ARGen            & \CT{VP-SDE}      & fix enc   &        &        & \XMark \\
LDLM~\tabcite{meshchaninov2026ldlm}            & \CT{VP-DDPM}     & learn enc &        &        & \XMark \\
Cola DLM~\tabcite{guo2026cola}\ARGen           & \CT{FM}          & learn enc &        &        & \XMark \\
\midrule
\multicolumn{6}{l}{\emph{Flow-based LMs}} \\
\midrule
CFM~\tabcite{roos2026categorical}              & \CT{FM}          & simplex   & \XMark &        &        \\
FLM~\tabcite{lee2026one}                      & \CT{FM}          & one-hot   & \XMark &        &        \\
DFM~\tabcite{potaptchik2026discrete}           & \CT{FM}          & simplex   & \XMark &        &        \\
LangFlow~\tabcite{chen2026langflow}            & \CT{Bregman FM}  & learn emb & \XMark &        &        \\
\midrule
\rowcolor[gray]{0.92}
\textbf{ELF (ours)}                            & \CT{\textbf{FM}} & \textbf{fix enc} &        &        &        \\
\bottomrule
\end{tabular}

\vspace{1em}

\begin{minipage}{0.75\textwidth}
\scriptsize
\setlength{\parskip}{0pt}
\setlength{\parindent}{0pt}
\noindent
\textsuperscript{\dag}\textit{Process:}
FM = Flow Matching~\cite{lipman2022flow,liu2022flow,albergo2025stochastic};
DDPM = Denoising Diffusion Probabilistic Model~\cite{ho2020denoising};
VP-DDPM/-SDE = variance-preserving DDPM / stochastic differential equation~\cite{song2020score};
Score-ODE = probability-flow ODE~\cite{song2020score};
SDE/DDIM = continuous-time SDE~\cite{song2020score} integrated with the deterministic DDIM solver;
VLB = variational lower bound, specifically Plaid's $T\!\to\!\infty$ continuous-time limit~\cite{gulrajani2023likelihood};
RDM = Riemannian Diffusion Mixture, applied to the categorical sphere by RDLM~\cite{jo2025continuous};
Bregman FM = Flow Matching with a Bregman-divergence regression objective, used by LangFlow~\cite{chen2026langflow}. \\[0.4em]
\textsuperscript{\ddag}\textit{State:}
learn emb = jointly trained token embedding matrix;
fix emb = frozen pretrained embedding lookup;
fix enc = frozen pretrained encoder, optionally with a compressed autoencoder bottleneck on top;
learn enc = trainable latent encoder, such as one on top of frozen pretrained representations or a text VAE, trained jointly with the diffusion model;
simplex = vocabulary-shaped logit simplex or square-root simplex on the sphere;
one-hot = per-token one-hot stack over the vocabulary.
\end{minipage}

\vspace{-1.2em}
\end{table*}

\paragraph{Survey details.}
We provide a detailed survey in Tab.~\ref{tab:dlm-survey}. The survey summarizes representative continuous diffusion and flow-based language models along several design axes, including the underlying diffusion or flow process, the continuous state in which denoising is performed, whether intermediate denoising states are discretized during training or inference, and whether a separate decoder is required to map latent states back to text.

In particular, the \textit{Train per-step discr.} and \textit{Infer. per-step discr.} columns distinguish two different uses of intermediate discretization. \textit{Train per-step discr.} indicates that intermediate denoising states are mapped to token predictions during training and supervised with token-level objectives such as cross-entropy loss. This provides direct vocabulary-level guidance, but also couples intermediate denoising states to categorical predictions. \textit{Infer. per-step discr.} indicates that intermediate sampling states are explicitly projected back to token-aligned representations during generation, such as nearest-neighbor rounding in embedding space or argmax projection on a simplex. Methods without inference-time per-step discretization keep the sampling trajectory continuous and discretize only at the final step. The \textit{Sep. dec.} column indicates whether a method requires a separate decoder, trained either separately or jointly with the diffusion model, to map continuous latent representations back to discrete text.

\paragraph{Positioning of ELF.}
Tab.~\ref{tab:dlm-survey} shows that existing continuous DLMs differ substantially in where the denoising process is defined and how continuous states are mapped back to text. Many embedding-space and simplex-based methods use training-time per-step discretization through token-level objectives, commonly cross-entropy, at intermediate denoising steps. These objectives provide direct token-level guidance, while making the denoising trajectory more tightly coupled to vocabulary-level prediction. Latent Diffusion LMs often avoid such per-step vocabulary supervision, but often rely on DDPM-style or score-based formulations with DDPM noise schedules and require a distinct latent-to-text decoder, such as an autoregressive decoder, non-autoregressive decoder, or latent decompressor, to recover discrete tokens, trained either separately or jointly with the diffusion model.

ELF occupies a different design point. It formulates language generation as continuous-time Flow Matching in a frozen contextual embedding space and keeps the sampling trajectory continuous, applying discretization only at the final decoding step. Unlike prior latent Diffusion LMs, ELF does not require a separate decoder: a single shared-weight network performs intermediate denoising and recovers tokens at the final step through the unembedding layer.
\FloatBarrier
\section{Progressive Distillation of ELF for Few-Step Language Generation}
\label{sec:distillation}

We have shown that ELF achieves strong performance with 8--32 sampling steps, outperforming prior discrete and continuous DLMs that rely on distillation. However, its performance degrades as the number of sampling steps is further reduced, and one-/few-step generation remains challenging for ELF. Progressive distillation has been shown to substantially reduce the number of sampling steps while preserving generation quality~\citep{salimans2022progressive}. Motivated by this approach, we introduce \emph{ELF with progressive distillation} (ELF+PD), which distills a pretrained ELF teacher into a student model for few-step language generation.

\subsection{Method}

The key idea of progressive distillation~\citep{salimans2022progressive} is to compress $K$ teacher sampling steps into a single student step. Given a time interval $[t,r]$, we apply $K$ teacher sampling steps to move the noisy embedding from $\z_t$ to $\z_r$ using a numerical solver. We then convert the resulting teacher displacement into a new target:
\begin{equation}
     \tilde{\x} = \z_t + \frac{1 - t}{r - t} (\z_r - \z_t).
     \label{eq:pd_target}
\end{equation}

We train an ELF+PD student with parameters $\theta$ by minimizing
\begin{equation}
\distill = \mathbb{E}_{t, \x, \e} \big\| \x_{\theta}(\z_t, t) - \tilde{\x}\big\|^2 . \label{eq:distill_loss}
\end{equation}

We keep the two-branch training setup of ELF and use the same shared-weight denoiser and decoder. For the denoising branch, we replace the MSE loss in Eq.~(\ref{eq:loss}) with the distillation loss in Eq.~(\ref{eq:distill_loss}). For the decoding branch, we keep the original cross-entropy loss in Eq.~(\ref{eq:ce_loss}).

We progressively distill the teacher into a single-step student by halving the number of steps at each round as in~\citep{salimans2022progressive, sahoo2024simple, sahoo2025diffusion}. We perform five rounds of distillation using the curriculum shown in Tab.~\ref{tab:curriculum}. We use a fixed 64-step ELF teacher throughout all five rounds. The first 16-step student is initialized from the teacher, while each subsequent student is initialized from the student from the previous round. After the first round, each round halves the number of student steps. At a round targeting an $N$-step student, each student step matches $64/N$ teacher substeps, allowing the student to approximate the 64-step teacher trajectory using fewer sampling steps.

\begin{table}[t]
\centering
\caption{\textbf{ELF+PD distillation curriculum.} Each round halves the number of student steps while keeping a fixed 64-step teacher. Each step of an $N$-step student matches $64/N$ teacher substeps.}
\label{tab:curriculum}
\small
\begin{tabular}{ccc}
\toprule
\textbf{Round} & \textbf{Student steps $N$} & \textbf{Teacher substeps} \\
\midrule
$r_1$ & 16 & 4 \\
$r_2$ & 8 & 8 \\
$r_3$ & 4 & 16 \\
$r_4$ & 2 & 32 \\
$r_5$ & 1 & 64 \\
\bottomrule
\end{tabular}
\end{table}

\subsection{Experimental Setup}
We use the OpenWebText dataset and follow the experimental setup described in Sec.~\ref{sec:exp}. For each round of distillation, we train the model for one epoch using the same hyperparameter settings as in the original ELF training (see Tab.~\ref{tab:training_hparams}), with 0.1 warmup epoch. Similar to ELF, we use training-time CFG with self-conditioning: a CFG scale is sampled for each example, and the teacher and student are conditioned on the same self-conditioning CFG scale. We use the same logit-normal schedule to sample time steps for both the student model and the teacher model.

During inference, we use the SDE-inspired sampler with the same logit-normal time schedule used during training. For 1-, 2-, 4-, and 8-step generation, we set the noise reinjection scale to $\gamma=1.5$ and the self-conditioning CFG scale to $2.5$. For 16- and 32-step generation, we use $\gamma=2.0$ and a self-conditioning CFG scale of $2.0$.

\begin{figure}[t]
\centering
\includegraphics[width=1.0\linewidth]{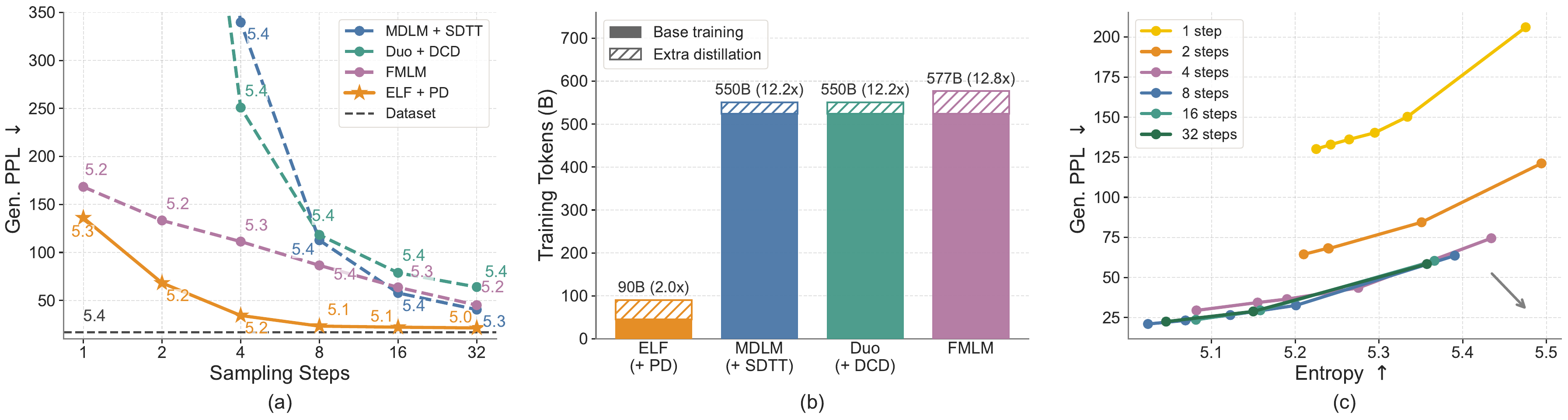}
\caption{\textbf{ELF with progressive distillation (ELF+PD) vs.\ distilled DLM baselines on OpenWebText.} \textbf{(a)} Generative perplexity against sampling steps: ELF+PD outperforms distilled baselines across all sampling steps. Per-point labels indicate entropy. \textbf{(b)}
Estimated training tokens: ELF+PD uses only $90$B tokens ($2.0\times$ the base model training), while other baselines require $550$--$577$B ($12\times$+) training tokens.
\textbf{(c)} Generative perplexity–entropy trade-off of ELF+PD across different numbers of sampling steps, obtained by sweeping the self-conditioning CFG scale. The trade-offs are comparable across 8–32 steps, while using fewer than 8 steps leads to degraded performance.
% from $0.5$ to $3.0$ (with $\gamma=1.5$).
}
\label{fig:distillation_combined}
\end{figure}

For the self-conditioning CFG sweep in Fig.~\ref{fig:distillation_combined}c, we sweep the self-conditioning CFG scale from $0.5$ to $3.0$ while fixing $\gamma=1.5$, and report only configurations with entropy greater than $5.0$ to exclude degenerate, repetitive text. To study the effect of curriculum distillation in Tab.~\ref{tab:per_round_results}, we fix $\gamma=1.5$ and the self-conditioning CFG scale to $2.5$.

\subsection{Results}
\paragraph{Few-step generation.}
Fig.~\ref{fig:distillation_combined}a compares ELF+PD with representative distilled discrete and continuous DLM baselines on OpenWebText: distilled versions of the discrete models (MDLM+SDTT~\citep{sahoo2024simple, deschenaux2024sdtt} and Duo+DCD~\citep{sahoo2025diffusion}), and the continuous flow-based model (FMLM~\citep{lee2026one}). ELF+PD achieves the lowest generative perplexity at every sampling budget while maintaining reasonable entropy. Benefiting from the data efficiency of ELF, ELF+PD also uses substantially fewer training tokens than the baselines, as shown in Fig.~\ref{fig:distillation_combined}b. Exact numbers are reported in Tab.~\ref{tab:few_step_owt}. We also compare the generative perplexity--entropy trade-off of ELF+PD under different numbers of sampling steps in Fig.~\ref{fig:distillation_combined}c. The trade-offs are comparable across 8--32 steps, while using fewer than 8 steps leads to degraded performance.

\begin{table}[t!]
\centering
\caption{\textbf{Few-step unconditional generation on OpenWebText.} We compare ELF+PD after five-round distillation with distilled discrete and continuous diffusion language model baselines. Baseline results are from Lee et al.~\cite{lee2026one} (Table 13 and Fig. 12).}
\label{tab:few_step_owt}
\scriptsize
\setlength{\tabcolsep}{3.4pt}
\renewcommand{\arraystretch}{1.0}
\begin{tabular}{ccccccccc}
\toprule
\multirow{2}{*}{\textbf{Steps}}
& \multicolumn{2}{c}{\textbf{MDLM + SDTT}~\citep{sahoo2024simple, deschenaux2024sdtt}}
& \multicolumn{2}{c}{\textbf{Duo + DCD}~\citep{sahoo2025diffusion}}
& \multicolumn{2}{c}{\textbf{FMLM}~\citep{lee2026one}}
& \multicolumn{2}{c}{\textbf{ELF + PD (Ours)}} \\
\cmidrule(lr){2-3}
\cmidrule(lr){4-5}
\cmidrule(lr){6-7}
\cmidrule(lr){8-9}
& \textbf{Gen. PPL} $\downarrow$ & \textbf{Entropy} $\uparrow$
& \textbf{Gen. PPL} $\downarrow$ & \textbf{Entropy} $\uparrow$
& \textbf{Gen. PPL} $\downarrow$ & \textbf{Entropy} $\uparrow$
& \textbf{Gen. PPL} $\downarrow$ & \textbf{Entropy} $\uparrow$ \\
\midrule
1 & 1260.86 & 5.26 & 5743.29 & 6.02 & 168.30 & 5.17 & \textbf{136.10} & 5.26 \\
2 & 877.22 & 5.34 & 891.16 & 5.41 & 133.29 & 5.25 & \textbf{68.25} & 5.24 \\
4 & 339.73 & 5.38 & 250.86 & 5.37 & 111.31 & 5.26 & \textbf{34.33} & 5.16 \\
8 & 112.66 & 5.41 & 118.21 & 5.41 & 86.50 & 5.36 & \textbf{23.18} & 5.07 \\
16 & 57.74 & 5.39 & 78.74 & 5.43 & 63.63 & 5.29 & \textbf{22.12} & 5.06 \\
32 & 40.41 & 5.34 & 63.98 & 5.40 & 45.09 & 5.25 & \textbf{21.32} & 5.04 \\
\bottomrule
\end{tabular}
\end{table}

\paragraph{Effect of distillation curriculum.}
As described above, we use a five-round curriculum to progressively distill the teacher into a one-step student. Each round halves the number of student steps and doubles the number of teacher substeps per interval. Tab.~\ref{tab:per_round_results} shows the student's performance across the curriculum. Early-round models perform well only with larger sampling budgets and collapse to degenerate outputs at smaller budgets, whereas later-round models substantially improve 1--4-step generation while maintaining reasonable entropy. After the final round, the one-step student achieves the best 1-, 2-, 4-, and 8-step generative perplexity; see qualitative examples in Sec.~\ref{sec:qualitative_examples_distill}.

\begin{table}[t!]
\centering
\caption{\textbf{Results across ELF with progressive distillation rounds and sampling steps.} Early-round students collapse at smaller sampling budgets, while later-round students substantially improve few-step performance. $^{*}$ indicates degenerate results, \ie, entropy below 5.0.}
\label{tab:per_round_results}
\scriptsize
\setlength{\tabcolsep}{4.2pt}
\renewcommand{\arraystretch}{1.0}
\begin{tabular}{ccccccccc}
\toprule
\multirow{2}{*}{\textbf{Round}}
& \multicolumn{2}{c}{\textbf{step = 1}}
& \multicolumn{2}{c}{\textbf{step = 2}}
& \multicolumn{2}{c}{\textbf{step = 4}}
& \multicolumn{2}{c}{\textbf{step = 8}} \\
\cmidrule(lr){2-3}
\cmidrule(lr){4-5}
\cmidrule(lr){6-7}
\cmidrule(lr){8-9}
& \textbf{Gen. PPL} $\downarrow$ & \textbf{Entropy} $\uparrow$
& \textbf{Gen. PPL} $\downarrow$ & \textbf{Entropy} $\uparrow$
& \textbf{Gen. PPL} $\downarrow$ & \textbf{Entropy} $\uparrow$
& \textbf{Gen. PPL} $\downarrow$ & \textbf{Entropy} $\uparrow$ \\
\midrule
$r_1$ & 4.9    & 1.72$^{*}$ & 119.4 & 5.10 & 143.4 & 5.37 & 58.0 & 5.26 \\
$r_2$ & 1.7    & 0.64$^{*}$ & 171.1 & 5.24 & 128.6 & 5.40 & 36.6 & 5.24 \\
$r_3$ & 30.7   & 3.60$^{*}$ & 153.1 & 5.44 & 69.0  & 5.37 & 27.9 & 5.20 \\
$r_4$ & 165.9        & 5.35 & 92.2  & 5.36 & 46.1  & 5.28 & 27.3 & 5.19 \\
$r_5$ & 136.1        & 5.26 & 68.2  & 5.24 & 34.3  & 5.16 & 23.2 & 5.07 \\
\bottomrule
\end{tabular}
\end{table}
\FloatBarrier
\section{Method Details}
\label{sec:method_details}

\begin{figure}
    \centering
    \includegraphics[width=1.0\linewidth]{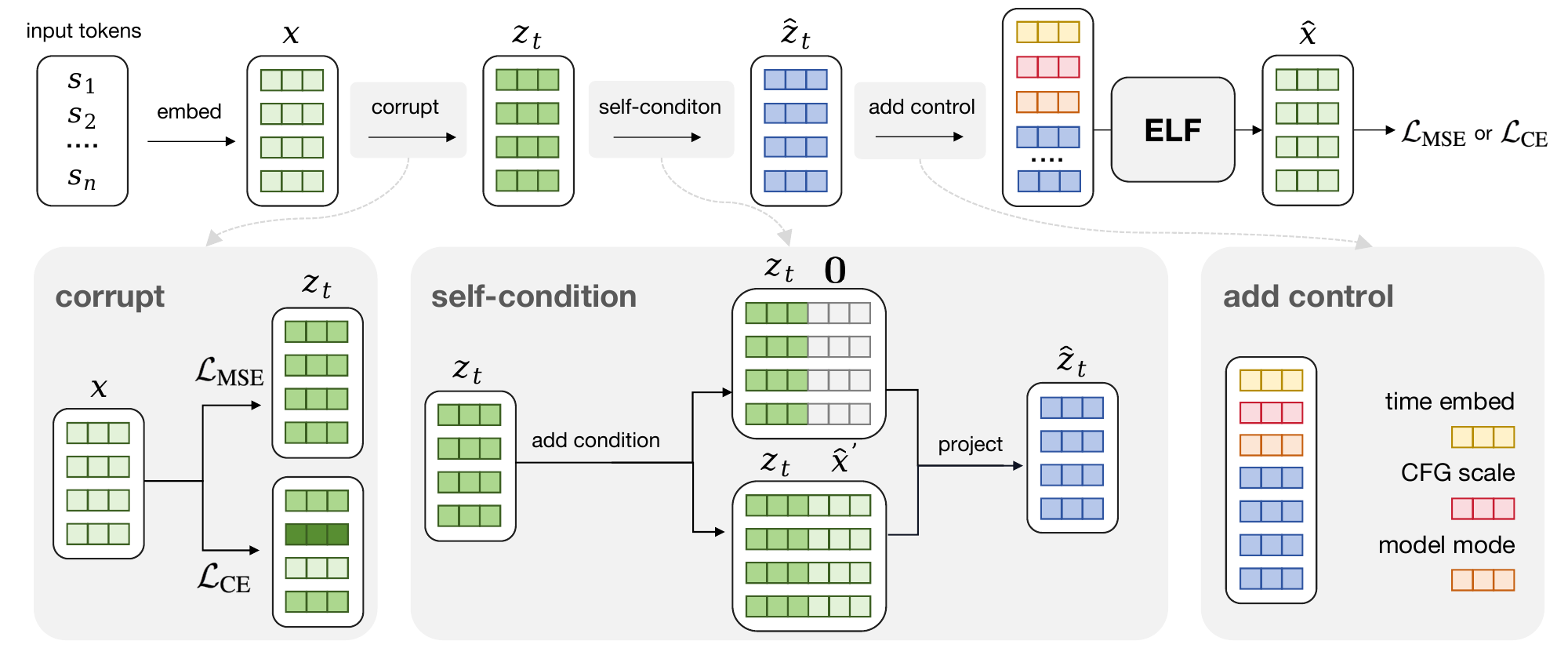}
    \caption{
    \textbf{Illustration of our training pipeline.} Starting from the clean embeddings $\x$, we apply different noise schedules in the two modes to obtain corrupted embeddings $\z_t$. We then apply self-conditioning by concatenating either $\mathbf{0}$ or the previous prediction $\hat{\x}'$ along the channel dimension, and project the concatenated embeddings back to the original dimension to form $\hat{\z}_t$. Next, we prepend control tokens to the embedding sequence, including time tokens in $[0,1]$, CFG scale tokens in $[0.5,5]$, and model-mode tokens indicating either denoising or decoding. The resulting sequence is fed into ELF to produce the final prediction $\hat{\x}$, which is supervised using either a denoising loss $\mse$ or a token-wise cross-entropy loss $\ce$. 
    }
\label{fig:method_details2}
\vspace{-1em}
\end{figure}

\subsection{Training}
\label{sec:app_train}

We show the full training pipeline in Fig.~\ref{fig:method_details2}. The input tokens are first encoded into clean embeddings $\x$, which then go through three key steps before being fed into the ELF model: corruption, self-conditioning, and adding control tokens for conditioning and guidance. In the denoising branch, the model predicts clean embeddings $\hat{\x}$ and is supervised with $\mse$. In the decoding branch, the same shared-weight network predicts embeddings that are then passed through an unembedding layer and supervised with $\ce$. The full training algorithm is shown in Alg.~\ref{alg:denoiser_cfg} and Alg.~\ref{alg:decoder_cfg}.

\paragraph{Embedding corruption.}
First, we corrupt the clean embeddings $\x$ by adding noise. Specifically, we use $\z_t = t \x + (1 - t)\e$ to obtain noisy embeddings $\z_t$, where $\e$ is Gaussian noise and $t$ is the time step. Before corruption, we first normalize the clean embeddings using the estimated mean and standard deviation from the OWT dataset. We use different noise schedules for different modes.

\lstdefinelanguage{PythonFuncColor}{
  language=Python,
  alsoletter={_},
  keywordstyle=\color{blue}\bfseries,
  commentstyle=\color{codeblue},
  stringstyle=\color{orange},
  showstringspaces=false,
  basicstyle=\ttfamily\small,
  literate=*
    {+}{{\color{codesign}+ }}{1}
    {*}{{\color{codesign}* }}{1}
    {/}{{\color{codesign}/ }}{1}
    {sample_per_token_p}{{\color{codefunc}sample\_per\_token\_p}}{1}
    {sample_t}{{\color{codefunc}sample\_t}}{1}
    {sample_sc_cfg_scale}{{\color{codefunc}sample\_sc\_cfg\_scale}}{1}
    {randn_like}{{\color{codefunc}randn\_like}}{1}
    {concat}{{\color{codefunc}concat}}{1}
    {encode}{{\color{codefunc}encode}}{1}
    {jvp}{{\color{codefunc}jvp}}{1}
    {stopgrad}{{\color{codefunc}stopgrad}}{1}
    {metric}{{\color{codefunc}metric}}{1}
    {def}{{\color{codefunc}def }}{1}
    {return}{{\color{codefunc}return }}{1}
    {mse_loss}{{\color{codefunc}mse\_loss}}{1}
    {ce_loss}{{\color{codefunc}ce\_loss}}{1}
    {argmax}{{\color{codefunc}argmax}}{1}
    {get_time_schedule}{{\color{codefunc}get\_time\_schedule}}{1}
    {unembed}{{\color{codefunc}unembed}}{1}
    {corrupt}{{\color{codefunc}corrupt}}{1}
    {random}{{\color{codefunc}random}}{1}
    {uniform}{{\color{codefunc}uniform}}{1}
    {where}{{\color{codefunc}where}}{1}
    {samplt_t}{{\color{codefunc}samplt\_t}}{1}
    {net}{{\color{codenetwork}net}}{1}
    {self_cond_proj}{{\color{codenetwork}self\_cond\_proj}}{1}
}

\lstset{
  language=PythonFuncColor,
  backgroundcolor=\color{white},
  basicstyle=\fontsize{9pt}{9.9pt}\ttfamily\selectfont,
  columns=fullflexible,
  breaklines=true,
  captionpos=b,
}

\begin{figure*}[t]\captionsetup{hypcap=false}% one anchor per algorithm (avoids duplicate PDF destinations)
\centering
\begin{minipage}[t]{\linewidth}
\begin{algorithm}[H]
\caption{{ELF} denoiser training with conditioning and guidance.}
\label{alg:denoiser_cfg}
\begin{lstlisting}[language=PythonFuncColor, escapechar=`]
# net(z, t, c, w, mode): ELF network with in-context conditioning
# self_cond_proj(z): Self-conditioning projection layer that converts concatenated embeddings back to the original embedding dimension
# self_cond_prob: Self-conditioning probability
# s: a sequence of discrete tokens
# c: condition (only for conditional generation)

x = encode(s)
t = sample_t()
w = sample_sc_cfg_scale()
e = randn_like(x)
z = t * x + (1 - t) * e
v = x - e

# z w/o self-conditioning
z_no_sc = self_cond_proj(concat([z, zeros_like(z)], dim=-1))
x_no_sc = net(z_no_sc, t, c, w, mode="denoise")
v_no_sc = (x_no_sc - z) / (1 - t)

# z w/ self-conditioning
z_sc = self_cond_proj(concat([z, stopgrad(x_no_sc)], dim=-1))
x_sc = net(z_sc, t, c, w, mode="denoise")
v_sc = (x_sc - z) / (1 - t)

# Compute CFG target
v_target = v + (1 - 1 / w) * (v_sc - v_no_sc) 

# Apply per-example self-conditioning mask
self_cond_mask = uniform(x.shape[0]) < self_cond_prob
v_pred = where(self_cond_mask, v_sc, v_no_sc)
v_target = where(self_cond_mask, v_target, v)
v_target = stopgrad(v_target)

# Compute v-loss
loss = mse_loss(v_pred, v_target)
\end{lstlisting}
\end{algorithm}
\end{minipage}
\end{figure*}

\begin{figure*}[t]\captionsetup{hypcap=false}% one anchor per algorithm (avoids duplicate PDF destinations)
\centering
\begin{minipage}[t]{\linewidth}
\begin{algorithm}[H]
\caption{{ELF} decoder training with conditioning and guidance.}
\label{alg:decoder_cfg}
\begin{lstlisting}[language=PythonFuncColor, escapechar=`]
# net(z, t, c, w, mode): ELF network with in-context conditioning
# self_cond_proj(z): Self-conditioning projection layer that converts concatenated embeddings back to the original embedding dimension
# s: a sequence of discrete tokens
# c: condition (only for conditional generation)

x = encode(s)
p = sample_per_token_p()
w = sample_sc_cfg_scale()
e = randn_like(x)
z = p * x + (1 - p) * e

# use z w/o self-conditioning
z = self_cond_proj(concat([z, zeros_like(z)], dim=-1))
h = net(z, t=1, c, w, mode="decode")
s_pred = unembed(h)
loss = ce_loss(s_pred, s)
\end{lstlisting}
\end{algorithm}
\end{minipage}

\end{figure*}

For the denoising branch, we sample the time step $t$ from a logit-normal distribution for each \textit{sequence}. Specifically, we draw $t' \sim \mathcal{N}(P_{\text{mean}}, P_{\text{std}}^2)$ and map it to the unit interval via $t = \sigma(t')$, where $\sigma(\cdot)$ denotes the sigmoid function. In all experiments, we use $P_{\text{mean}}=-1.5$ and $P_{\text{std}}=0.8$. 
We rescale the Gaussian noise by a factor of 2.

For the decoding branch, we train final-step discretization by conditioning the model on the decoder mode, \ie, $t=1$. At this time step, $\z_t$ corresponds to clean embeddings. Therefore, to make the final-step input nontrivial, we corrupt the clean embeddings with a per-token corruption level $p$ sampled from a different noise schedule. Specifically, we draw $p$ from a logit-normal distribution with $P_{\text{mean}}=0.8$ and $P_{\text{std}}=0.8$, and form $\tilde{\z}=p\x+(1-p)\e$, multiplying $\e$ by a noise scale. We use noise scales of 5 and 1 for OWT and conditional generation tasks, respectively. As a result, the corruption level varies across tokens within the same sequence. This design encourages the shared-weight decoder mode to recover corrupted embeddings from their surrounding context, making final-step discretization more robust to imperfect embeddings produced by the denoiser at inference time.

\paragraph{Self-conditioning.}

We apply self-conditioning following prior work~\citep{chen2022analog}. During training, with a certain probability, we perform an additional forward pass to obtain the predicted embeddings $\hat{\x}'$, which are concatenated with the noisy embeddings $\z_t$ along the channel dimension. We stop the gradient through the predicted embeddings $\hat{\x}'$. For the remaining examples, we concatenate $\z_t$ with all-zero embeddings $\mathbf{0}$ instead. Since this concatenation doubles the channel dimension, we project it back to the original dimension using a linear layer. We apply self-conditioning with $\hat{\x}'$ in the denoising branch with 50\% probability. For the decoding branch, we always use $\mathbf{0}$ as the self-conditioning input, as shown in Alg.~\ref{alg:decoder_cfg}.

\paragraph{Training-time CFG.}
As discussed in Sec.~\ref{sec:control}, our model performs training-time CFG~\citep{geng2025mean,geng2025improved,chen2025visual,tang2025diffusion} with self-conditioning. 
In training-time CFG, the network is designed to model the post-combination quantity $\v^\textrm{cfg}_\theta$, rather than the pre-combination quantity $\v_\theta$.
Following~\citep{geng2025mean,geng2025improved}, the regression target $\v_{\textrm{target}}$ is now:
\begin{equation}
\v_{\textrm{target}} = \x - \e + \left(1 - \frac{1}{\omega}\right) \bigl(\v^\textrm{cfg}_\theta(\z_t \mid t, \cond, \omega) - \v^\textrm{cfg}_\theta(\z_t \mid t, \varnothing, \omega)\bigr),
\label{eq:guided_v}
\end{equation}
where $\omega$ is the guidance scale. When $\omega=1$, this reduces to the case without training-time CFG. 
The training loss is $\| \v^\textrm{cfg}_\theta(\cdot) - \v_{\textrm{target}} \|^2$~\citep{geng2025mean,geng2025improved}. See Alg.~\ref{alg:denoiser_cfg}.
For each training example, we randomly sample a self-conditioning CFG scale $\omega \in [0.5, 5.0]$ from a power distribution biased toward smaller values~\citep{geng2025mean,geng2025improved}. Since ELF uses $\x$-prediction, the quantity $\v$ is always converted from its $\x$ prediction counterpart (conditional or unconditional).

Our model uses a diverse set of conditions. Standard diffusion models typically implement conditioning through adaLN-Zero~\citep{peebles2023scalable}, which combines all conditioning signals through summation. This design becomes less effective when many heterogeneous conditions are present. Therefore, we adopt in-context conditioning~\citep{geng2025improved} by prepending a set of \textit{control} tokens that encode the conditioning information. Each control-token embedding has the same dimensionality as a standard language-token embedding. We prepend three types of control tokens: 4 time tokens with values in $[0,1]$, 4 CFG-scale tokens sampled from $[0.5,5]$, and 4 model-mode tokens indicating either denoising or decoding. These tokens are jointly trained with the model. All continuous values, \ie, time and CFG scale, are encoded with positional embeddings.

For conditional generation, we place the clean embeddings of the conditioning sequence immediately after the control tokens and before the target sequence to be generated. The model then performs bidirectional self-attention over the concatenated sequence of conditioning and target tokens. The conditioning embeddings are kept uncorrupted during training. To enable CFG for conditional generation, we randomly drop the condition with 10\% probability by zeroing out the embeddings of the conditioning sequence. This allows the model to learn both conditional and unconditional generation under the same framework.

\subsection{Inference}
\label{sec:app_inference}

We show the full inference algorithm in Alg.~\ref{alg:inference_cfg}. Since the self-conditioning CFG scale is provided through in-context conditioning, changing $\omega$ does not require an additional inference pass. By modifying $\omega$ as a model input, we can flexibly control the trade-off between generation quality and diversity.

\lstdefinelanguage{PythonFuncColor}{
  language=Python,
  alsoletter={_},
  keywordstyle=\color{blue}\bfseries,
  commentstyle=\color{codeblue},
  stringstyle=\color{orange},
  showstringspaces=false,
  basicstyle=\ttfamily\small,
  literate=*
    {+}{{\color{codesign}+ }}{1}
    {*}{{\color{codesign}* }}{1}
    {/}{{\color{codesign}/ }}{1}
    {sample_t}{{\color{codefunc}sample\_t}}{1}
    {sample_cfg_scale}{{\color{codefunc}sample\_cfg\_scale}}{1}
    {randn}{{\color{codefunc}randn}}{1}
    {randn_like}{{\color{codefunc}randn\_like}}{1}
    {concat}{{\color{codefunc}concat}}{1}
    {jvp}{{\color{codefunc}jvp}}{1}
    {stopgrad}{{\color{codefunc}stopgrad}}{1}
    {metric}{{\color{codefunc}metric}}{1}
    {def}{{\color{blue}def }}{1}
    {sde_step}{{\color{codefunc}sde\_step}}{1}
    {ode_step}{{\color{codefunc}ode\_step}}{1}
    {return}{{\color{codefunc}return }}{1}
    {mse_loss}{{\color{codefunc}mse\_loss}}{1}
    {ce_loss}{{\color{codefunc}ce\_loss}}{1}
    {argmax}{{\color{codefunc}argmax}}{1}
    {get_time_schedule}{{\color{codefunc}get\_time\_schedule}}{1}
    {unembed}{{\color{codefunc}unembed}}{1}
    {corrupt}{{\color{codefunc}corrupt}}{1}
    {zeros_like}{{\color{codefunc}zeros\_like}}{1}
    {zeros}{{\color{codefunc}zeros}}{1}
    {uniform}{{\color{codefunc}uniform}}{1}
    {samplt_t}{{\color{codefunc}samplt\_t}}{1}
    {net}{{\color{codenetwork}net}}{1}
    {self_cond_proj}{{\color{codenetwork}self\_cond\_proj}}{1}
}

\lstset{
  language=PythonFuncColor,
  backgroundcolor=\color{white},
  basicstyle=\fontsize{9pt}{9.9pt}\ttfamily\selectfont,
  columns=fullflexible,
  breaklines=true,
  captionpos=b,
}

\begin{figure*}[t!]\captionsetup{hypcap=false}% one anchor per algorithm (avoids duplicate PDF destinations)
\centering
\begin{minipage}[t]{\linewidth}
\begin{algorithm}[H]
\caption{{ELF} inference with conditioning and guidance.}
\label{alg:inference_cfg}
\begin{lstlisting}[language=PythonFuncColor, escapechar=`]
# net(z, t, c, w, mode): ELF network with in-context conditioning
# self_cond_proj(z): Self-conditioning projection layer that converts concatenated embeddings back to the original embedding dimension
# shape: embeddings shape
# ts: discretized time grid over [0, 1] with N intervals
# c: condition (only for conditional generation)
# w: self-conditioning CFG scale

z = randn(shape)
x_pred = zeros(shape)

for i in range(len(ts) - 1):
    t = ts[i]
    dt = ts[i + 1] - ts[i]
    # Self-condition on the previous prediction
    z_sc = self_cond_proj(concat([z, x_pred], dim=-1))
    x_pred = net(z_sc, t, c, w, mode="denoise")
    # convert x prediction to velocity
    v = (x_pred - z) / (1 - t)
    z = z + dt * v

# decoding
z = self_cond_proj(concat([z, zeros_like(z)], dim=-1))
h = net(z, t=1, c, w, mode="decode")
# unembedding
token_logits = unembed(h)
tokens = argmax(token_logits)
\end{lstlisting}
\end{algorithm}
\end{minipage}

\end{figure*}

\begin{figure*}[t!]\captionsetup{hypcap=false}% one anchor per algorithm (avoids duplicate PDF destinations)
\centering
\begin{minipage}[t]{\linewidth}
\begin{algorithm}[H]
\caption{{ELF} inference with different samplers.}
\label{alg:sampler}
\begin{lstlisting}[language=PythonFuncColor, escapechar=`]
# z: noisy embeddings of current time step
# t: current time step
# dt: time interval, t_next - t
# gamma: controls the amount of noise added back in the SDE sampler

def ode_step(z, t, dt):
    x_hat = net(z, t, mode="denoise")
    v = (x_hat - z) / (1 - t)
    z =  z + dt * v
    return z

def sde_step(z, t, dt, gamma):
    # Re-inject noise and move back to the corresponding time step
    # The jump size is defined relative to the time-step interval
    e = randn_like(z)
    alpha = 1 - gamma * dt
    t_back = alpha * t
    z_back = alpha * z + (1 - alpha) * e

    x_hat = net(z_back, t_back, mode="denoise")
    v = (x_hat - z_back) / (1 - t_back)
    z =  z + dt * v
    return z


\end{lstlisting}
\end{algorithm}
\end{minipage}
\end{figure*}

\paragraph{Time schedule.}
We discretize the continuous time interval $t \in [0, 1]$ into $T$ intervals using a logit-normal time schedule. Specifically, we sample $T-1$ time steps from the same logit-normal distribution used for the denoising branch during training and sort them to form the intermediate points. We use $P_{\text{mean}}=-1.5$ and $P_{\text{std}}=0.8$ to match the training-time logit-normal distribution. We ensure that the first interval starts at $t=0$ and the last interval ends at $t=1$. This schedule produces smaller intervals when $t$ is close to 0 and larger intervals as $t$ approaches 1. It shows strong empirical performance, likely because the noisier regime requires finer discretization and the schedule better matches the noise distribution used during training.

\vspace{-0.5em}
\paragraph{Samplers.}

Our method supports both deterministic ODE sampling and an SDE-inspired stochastic sampler. The main algorithm in Alg.~\ref{alg:dlm-inference} uses the ODE sampler for simplicity, while Alg.~\ref{alg:sampler} summarizes one-step updates for both samplers.

The SDE variant is motivated by the SDE associated with Flow Matching~\citep{ma2024sit}, whose dynamics can be interpreted as injecting infinitesimal noise at each step. In practice, we adopt a simple approximation that re-injects Gaussian noise at each sampling step while shifting the time variable slightly toward the noise regime. We introduce a noise re-injection scale $\gamma$ to control the amount of stochasticity added at each step. The denoiser is then evaluated on this perturbed state, and its clean-embedding prediction is used to update the original state. When $\gamma=0$, no stochastic perturbation is applied, and the update reduces to deterministic ODE sampling.

\vspace{-0.5em}
\paragraph{CFG for conditional generation.} 
We apply standard CFG by combining the conditional and unconditional predictions. Similarly, we use the CFG scale to control the guidance strength.
\FloatBarrier
% \section{Additional System-Level Comparison}
% \label{sec:app_tradeoff}

% We compare the generative perplexity--entropy trade-off of ELF-B with those of other continuous and discrete baselines at the same scale (Fig.~\ref{fig:tradeoff}a), and that of its distilled counterpart, ELF+PD, with those of other distilled baselines (Fig.~\ref{fig:tradeoff}b). We use results from Fig.~\ref{fig:system_level_comparison}a for the undistilled models and from Fig.~\ref{fig:distillation_combined}a for the distilled models. Each point corresponds to a single generation setting with a different number of sampling steps. A frontier closer to the lower-right corner represents a better trade-off, indicating lower generative perplexity and higher mean unigram entropy. ELF and its distilled version can extend into a lower-generative-perplexity regime while maintaining reasonable entropy.

% \begin{figure}[t!]
% \centering
% \includegraphics[width=0.75\linewidth]{figures/tradeoff_panels.pdf}
% \caption{\textbf{System-level comparison on the generative perplexity--entropy trade-off.} Each point is obtained by varying the number of sampling steps. \textbf{(a)} We compare ELF-B with discrete and continuous DLM baselines without distillation. \textbf{(b)} We compare ELF+PD, the distilled version of ELF-B, with other distilled baselines.}
% \label{fig:tradeoff}
% \end{figure}
\FloatBarrier

\section{Additional Generation Quality Evaluations}
\label{sec:additional_generation_quality}

\subsection{MAUVE}
\label{sec:mauve}

We evaluate ELF-B using MAUVE~\citep{pillutla2021mauve} with a ModernBERT-large featurizer~\citep{warner2025modernbert}, as in prior work~\citep{yang2026replaid}. MAUVE measures the distributional similarity between generated text and the original dataset, complementing our main evaluation metrics of generative perplexity and entropy. We use a 512-token truncation length and compare approximately 5,120 generations from each method against the full reference set of one million OpenWebText blocks. For each sampling budget, we report the ELF-B configuration that achieves the highest MAUVE score.

\begin{figure}[h]
    \centering
    \includegraphics[width=0.4\linewidth]{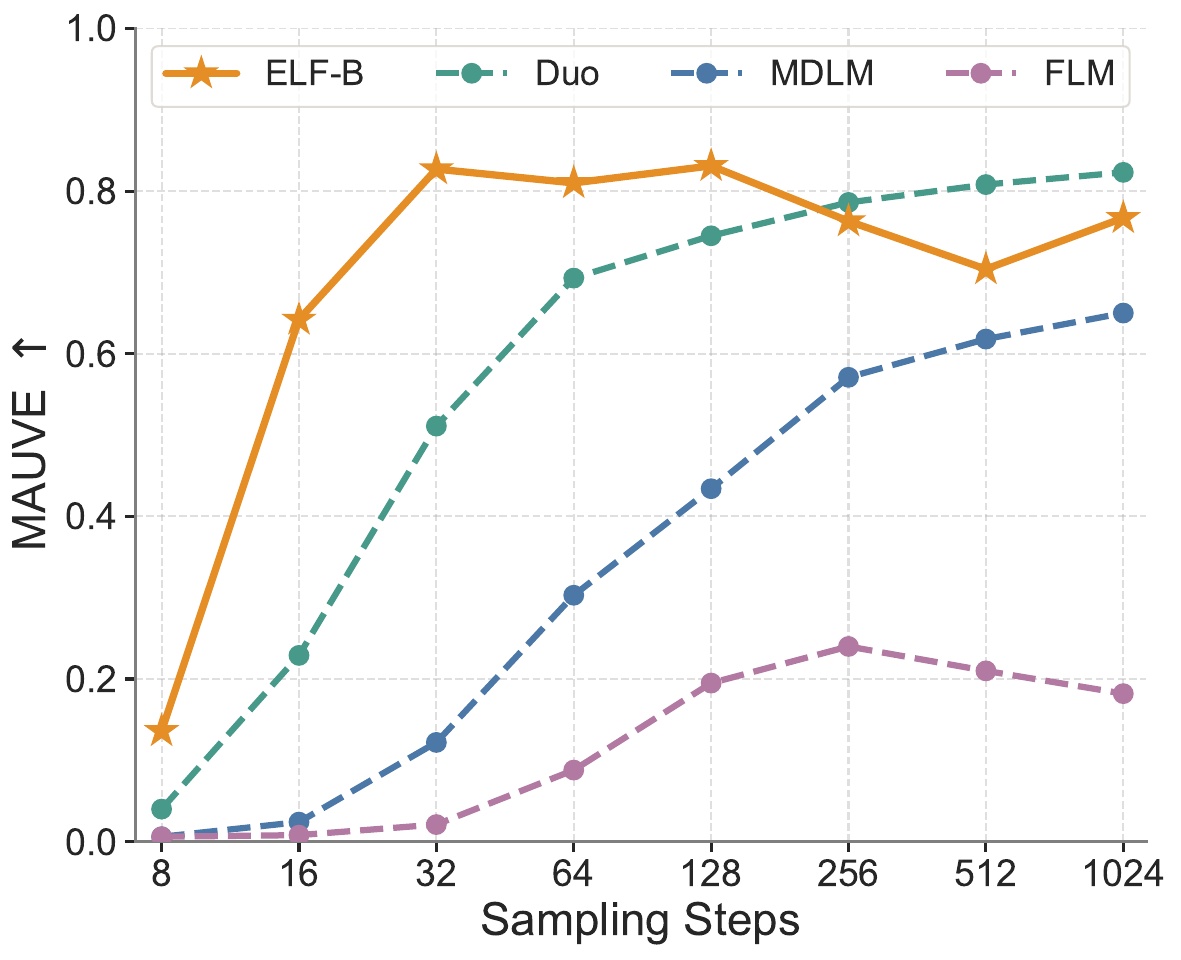}
    \caption{\textbf{MAUVE on OpenWebText across sampling budgets.} ELF-B achieves strong performance, especially in the low-to-moderate sampling-step regime.}
    \label{fig:mauve}
\end{figure}

As shown in Fig.~\ref{fig:mauve}, ELF-B performs particularly well in the low-to-moderate sampling-step regime, outperforming all listed baselines from 8 to 128 steps and reaching a MAUVE score of 0.827 with only 32 steps. At 128 steps, it achieves its highest score of 0.831. These results demonstrate that ELF-B achieves strong distributional generation quality with substantially fewer sampling steps.

\subsection{LLM-as-a-Judge Evaluation}
\label{sec:llm_judge}

We additionally conduct a per-sample LLM-as-a-judge evaluation as a proxy for human evaluation. Claude Opus 4.8 evaluates each unconditional generation independently on a 1--5 scale without access to the model identity.

We consider five dimensions. \textit{Fluency} measures grammaticality and sentence formation; \textit{non-repetition} measures the absence of degenerate loops and redundant content; \textit{cleanliness} measures the absence of artifacts such as broken subwords, corrupted characters, and stray tokens; \textit{coherence} measures connections between adjacent sentences and consistency of the overall topic; and \textit{sensibility} measures whether the content is meaningful and internally consistent rather than merely grammatical. We also report a separate holistic \textit{overall} score that assesses the overall generation quality.

We use the best-performing configuration for each model in the system-level comparison. ELF-B uses 32 sampling steps with a self-conditioning CFG scale of 3 and the SDE sampler, whereas Duo, MDLM, and FLM use 1,024 sampling steps. We evaluate 500 samples for each method. We bold the best score in each dimension and scores that are not significantly different from it ($p > 0.05$).

As shown in Tab.~\ref{tab:llm_judge}, the holistic overall scores are similar across all methods (1.91--1.97), suggesting broadly comparable generation quality. ELF-B achieves the highest scores in fluency (3.43), cleanliness (3.86), and coherence (2.05), indicating more well-formed text with fewer generation artifacts. In contrast, Duo performs better on non-repetition and sensibility.

\begin{table}[H]
    \centering
    \caption{\textbf{LLM-as-a-judge scores on unconditional generations.} Scores range from 1 to 5, with higher values indicating better quality. Bold indicates the best score in each dimension and scores that are not significantly different from it ($p>0.05$).}
    \label{tab:llm_judge}
    \scalebox{0.8}{
    \begin{tabular}{lcccccc}
        \toprule
        \textbf{Method} & \textbf{Fluency} & \textbf{Non-repetition} & \textbf{Cleanliness} & \textbf{Coherence} & \textbf{Sensibility} & \textbf{Overall} \\
        \midrule
        Duo~\citep{sahoo2025diffusion} & 2.99 & \textbf{3.89} & 3.68 & 2.01 & \textbf{1.75} & \textbf{1.96} \\
        MDLM~\citep{sahoo2024simple} & 2.82 & 3.83 & 3.28 & 1.98 & 1.54 & 1.94 \\
        FLM~\citep{lee2026one} & 2.34 & 3.38 & 2.77 & 1.90 & 1.45 & 1.91 \\
        ELF-B & \textbf{3.43} & 2.40 & \textbf{3.86} & \textbf{2.05} & 1.58 & \textbf{1.97} \\
        \bottomrule
    \end{tabular}
    }
\end{table}

\FloatBarrier
\section{Additional Ablations}

\label{sec:additional_ablations}

In this section, we present additional ablations of our design choices. Unless otherwise specified, all experiments use the logit-normal time schedule with either a 64-step ODE sampler or a 64-step SDE sampler with $\gamma=1$. As before, we evaluate the generative perplexity--entropy trade-off by varying the self-conditioning CFG scale. We use red to indicate regions with poor generation quality, \ie, entropy below 5.0, which often corresponds to repetitive or degenerate sentences, or generative perplexity above 300, which often corresponds to semantically meaningless or ungrammatical sentences. All models are trained for the same number of steps, with all other configurations kept the same as the default setting.
\subsection{Prediction Targets}
\label{sec:prediction_type}
\begin{figure}[t]
    \centering
    \includegraphics[width=1.0\linewidth]{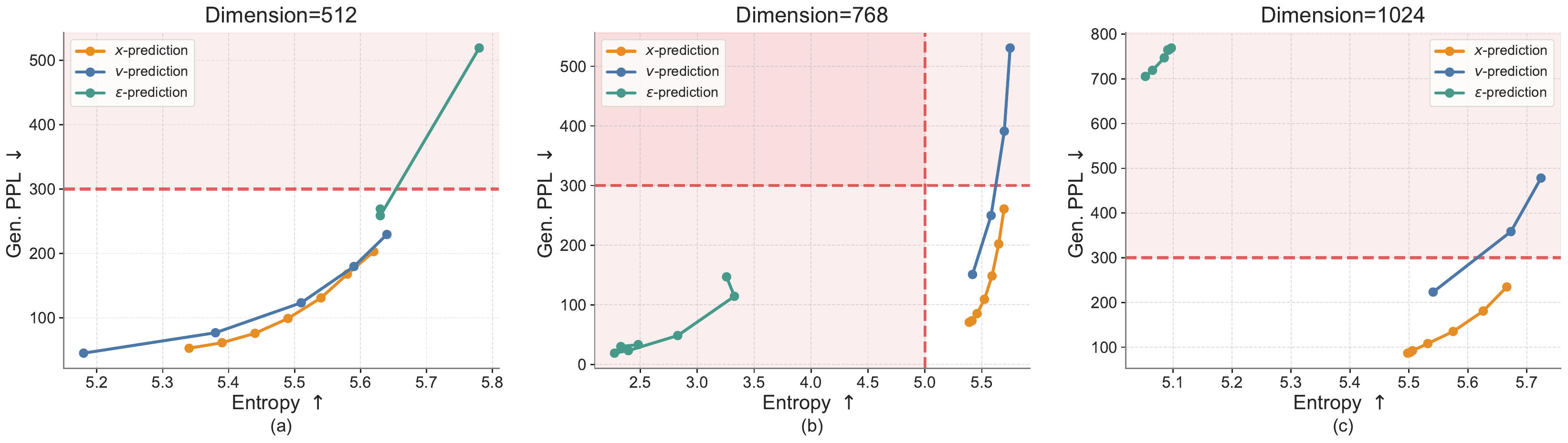}
    \caption{
    \textbf{Effects of prediction targets.} We vary the input dimension from 512 to 768 and 1024 by using T5-small, T5-base, and T5-large encoders, respectively. Across all input dimensions, $\x$-prediction remains stable and performs well. In contrast, $\v$-prediction performs well at 512 dimensions but degrades at higher dimensions, while $\e$-prediction collapses across all dimensions from 512 to 1024. The red region indicates poor-quality generations, where entropy falls below 5 (\eg, repetitive sentences) or generative perplexity exceeds 300 (\eg, meaningless or ungrammatical sentences). This aligns with the hypothesis from prior work that high-dimensional clean data often lies on a low-dimensional manifold~\citep{li2025back}.
    }
    \label{fig:prediction_type}
\end{figure}

Our model directly predicts the clean embeddings $\x$ ($\x$-prediction). This allows us to use a unified denoiser and decoder through weight sharing and jointly optimize the model with both the denoising objective $\mse$ and the token-level objective $\ce$. Prior work has also suggested that $\x$-prediction is essential, as high-dimensional clean data tends to lie on a low-dimensional manifold~\citep{li2025back}.

Here, we further study the effect of prediction targets. Specifically, since there are three quantities and two constraints: linear interpolation $\z_t = t \, \x + (1 - t)\, \e$ and flow velocity $\v = \x - \e$, the network can be trained to predict one of these quantities, \ie, $\x$-, $\v$-, or $\e$-prediction. To study this in a controlled setting, we use a two-stage pretrained encoder-decoder setup: a pretrained T5 encoder maps tokens into continuous embeddings, and a decoder is trained to reconstruct masked and noisy embeddings (see~\Cref{sec:ablation_studies_details} for details). We train only the denoising model while keeping both the encoder and decoder fixed. We use adaLN-Zero conditioning and a 64-step ODE sampler to plot the generative perplexity--entropy trade-off curve.

\begin{figure}[t]
    \centering
    \includegraphics[width=0.75\linewidth]{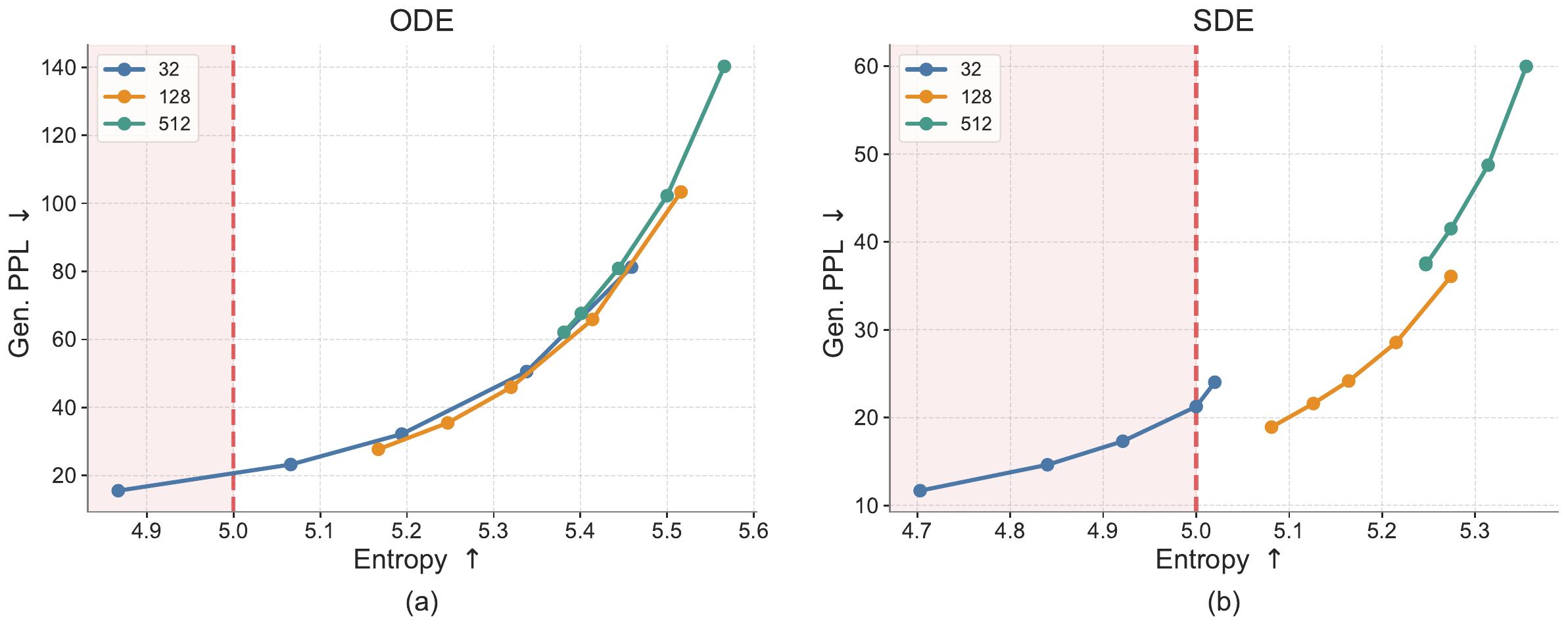}
    \caption{\textbf{Effect of bottleneck dimension.}
    We compare bottleneck dimensions of 32, 128, and 512 under ODE and SDE sampling. A moderate bottleneck dimension of 128 provides the best generative perplexity--entropy trade-off, while overly small or large bottlenecks either reduce diversity or hurt generative perplexity. Red indicates regions with poor generation quality, \ie, entropy below 5.}
    \label{fig:bottleneck}
\end{figure}

To study how prediction targets behave as the embedding dimension increases, we consider T5-small, T5-base, and T5-large encoders, corresponding to embedding dimensions of 512, 768, and 1024, respectively. We set the bottleneck dimension equal to the corresponding input embedding dimension. As shown in Fig.~\ref{fig:prediction_type}, $\x$-prediction remains the most stable across all dimensions, maintaining a reasonable generative perplexity--entropy trade-off even at 1024 dimensions. In contrast, $\v$-prediction is competitive at 512 dimensions but degrades as the dimension increases, with substantially higher generative perplexity at 768 and 1024 dimensions. $\e$-prediction collapses across all dimensions, either achieving extremely low entropy or high generative perplexity, indicating repetitive, degenerate, or ungrammatical generations. These results support the hypothesis that clean-data prediction is better suited to high-dimensional language representations, consistent with findings from prior work~\citep{li2025back}.

\subsection{Bottleneck}
\label{sec:bottleneck}
Our model uses a bottleneck design that projects encoder representations into a lower-dimensional space before mapping them back to the model hidden size.
This design is motivated by the hypothesis that natural data may lie on a low-dimensional manifold within the high-dimensional embedding space. 
We compare bottleneck dimensions of 32, 128, and 512, and show the results in Fig.~\ref{fig:bottleneck}.
The bottleneck dimension has a clear effect on the generative perplexity--entropy trade-off. Under ODE sampling, all three bottleneck sizes follow a similar frontier, but smaller bottlenecks tend to reach lower generative perplexity at the cost of lower entropy. Under SDE sampling, the differences become more significant: the 32-dimensional bottleneck achieves the lowest generative perplexity but often lies in the low-entropy region, indicating reduced diversity, whereas the 512-dimensional bottleneck maintains higher entropy but suffers from substantially worse generative perplexity. The 128-dimensional bottleneck provides the best overall balance, achieving strong generative perplexity while preserving reasonable entropy. We therefore use a bottleneck dimension of 128 as the default setting. This finding is also consistent with prior work~\citep{li2025back}, which observes that an appropriate bottleneck can improve performance.

\begin{figure}
    \centering
    \includegraphics[width=0.75\linewidth]{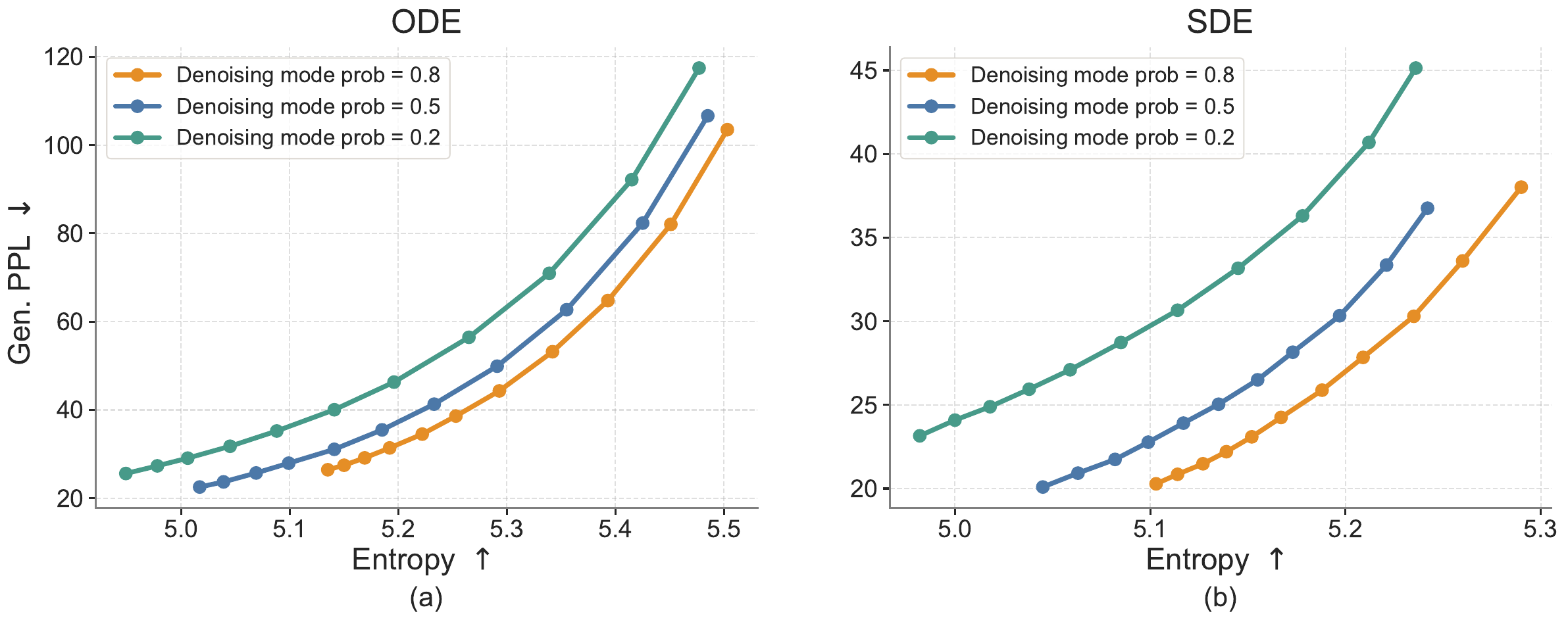}
    \caption{
    \textbf{Effect of the denoising mode probability during training.} 
    This probability controls the allocation between denoising and decoding updates in the shared-weight denoiser-decoder model. A denoising mode probability of $0.8$ provides the best generative perplexity--entropy trade-off across both ODE and SDE samplers.
    }
\label{fig:denoiser_mode_ratio}
\end{figure}

\begin{figure}[t]
    \centering
    \begin{minipage}[t]{0.48\linewidth}
        \centering
        \includegraphics[width=\linewidth]{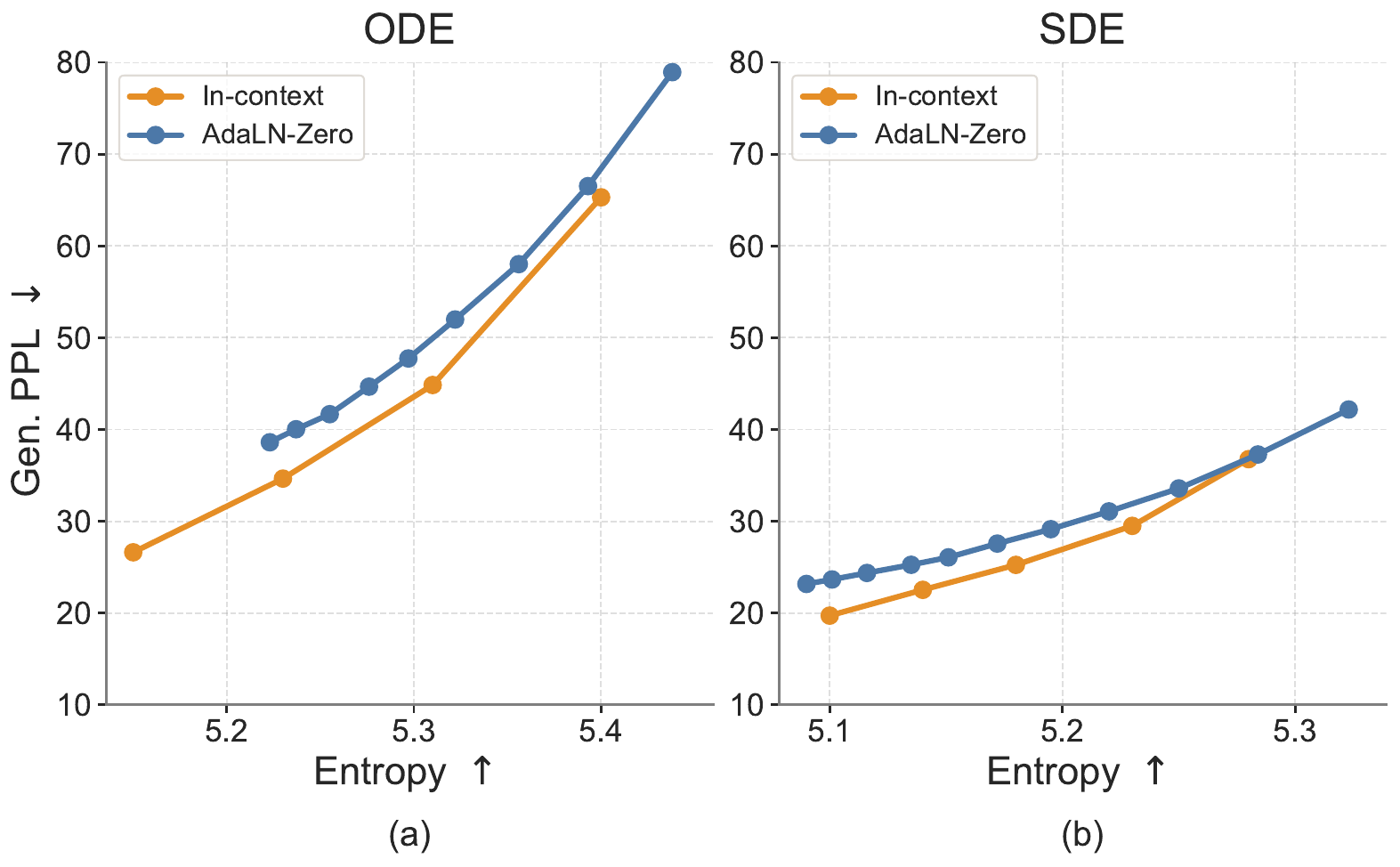}
        \caption{
        \textbf{Effect of conditioning strategies.}
        We compare in-context conditioning with adaLN-Zero conditioning. In-context conditioning slightly improves performance while substantially reducing the number of model parameters.
        }
        \label{fig:condition_choice}
    \end{minipage}
    \hfill
    \begin{minipage}[t]{0.48\linewidth}
        \centering
        \includegraphics[width=\linewidth]{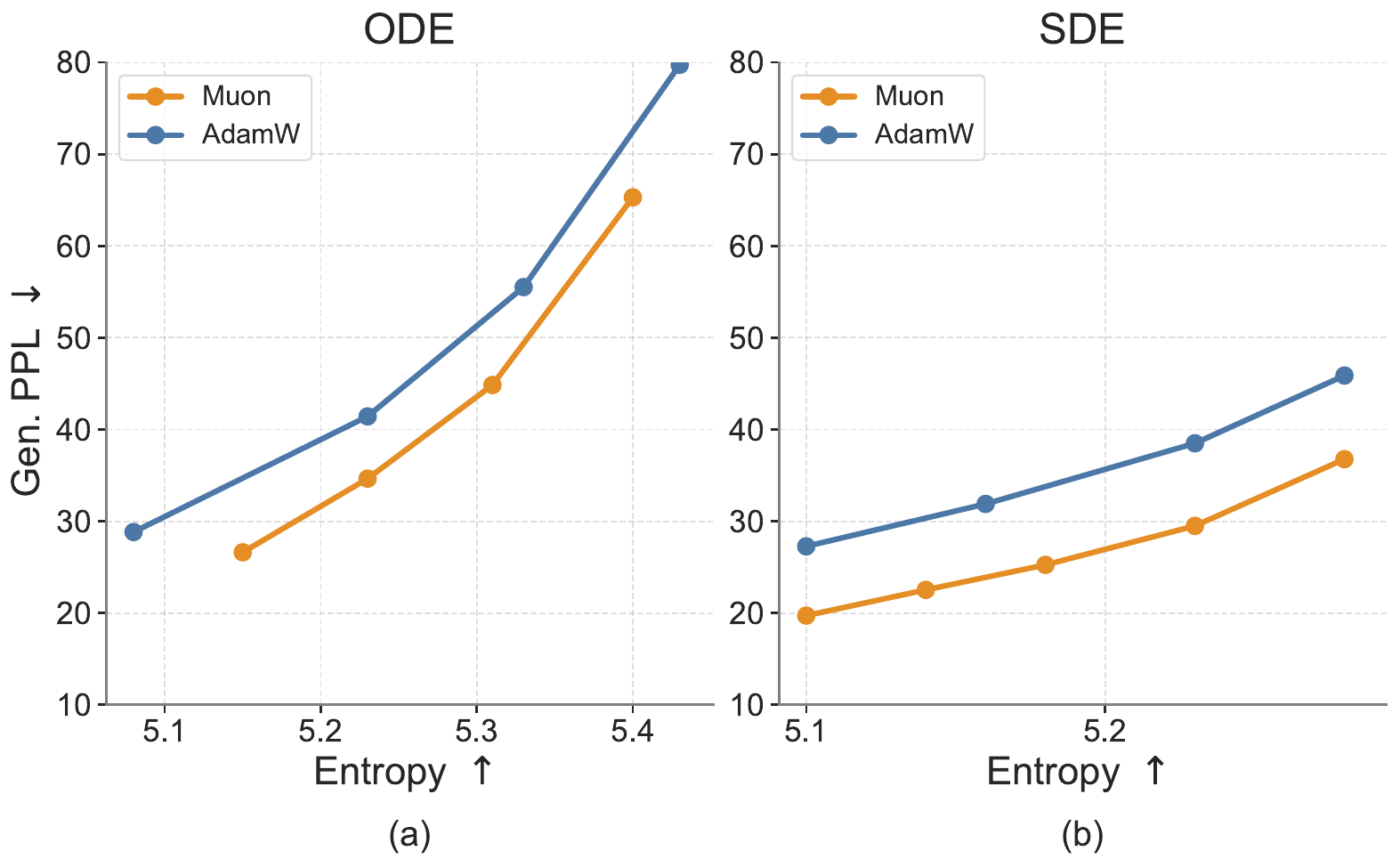}
        \caption{
        \textbf{Effect of optimizers.}
        We compare generation quality under different optimizers using Muon and AdamW. 
        Muon achieves lower generative perplexity at comparable entropy under both ODE and SDE sampling methods. 
        }
        \label{fig:optimizer_choice}
    \end{minipage}
\end{figure}

\subsection{Denoising Mode Probability}
\label{sec:denoiser_mode_ratio}

Since ELF is trained with both MSE and CE losses through a shared-weight denoiser-decoder, each training step is assigned to either denoising mode or decoding mode. The denoising-mode probability controls this allocation: a higher probability emphasizes learning the continuous denoising dynamics, while a lower probability provides more supervision for mapping embeddings back to tokens. We study this trade-off by varying the denoising-mode probability during training.

As shown in Fig.~\ref{fig:denoiser_mode_ratio}, assigning a low probability to the denoising mode consistently degrades the generative perplexity--entropy trade-off, especially under SDE sampling. This suggests that the model requires sufficient training on the denoising process. Among the configurations tested, a denoising mode probability of $0.8$ achieves the best overall trade-off across both ODE and SDE samplers. We therefore use $0.8$ as the default denoising mode probability in our main experiments.

\subsection{Conditioning Strategies}

\label{sec:condition_strategy}

As discussed in Sec.~\ref{sec:control}, our model is conditioned on the time step, CFG scale, and model mode. We use in-context conditioning for these signals by prepending them as condition tokens to the input sequence, allowing the model to attend to them through full attention. This differs from the conventional adaLN-Zero conditioning design, which typically introduces additional model components to process the conditioning inputs. We compare these two designs in Fig.~\ref{fig:condition_choice}. In-context conditioning performs slightly better while avoiding the substantial parameter overhead introduced by adaLN-Zero (ELF-B's parameter count is reduced from 148M to 105M). Therefore, we use in-context conditioning as our default setting.

\subsection{Optimizers}
\label{sec:optimizer}
We evaluate the impact of optimizer choice, comparing Muon~\citep{keller2024muon} and AdamW~\citep{loshchilov2017decoupled}, and show the results in Fig.~\ref{fig:optimizer_choice}. We tune the hyperparameters for both optimizers to obtain their best performance: for Muon, we use a learning rate of $2\times10^{-3}$; for AdamW, we use a learning rate of $1\times10^{-4}$ with $\beta_1=0.9$ and $\beta_2=0.95$. During training, Muon achieves lower loss within the same number of steps. During inference, models trained with Muon consistently achieve a better generative perplexity--entropy trade-off than those trained with AdamW under both samplers. The improvement is especially significant under SDE sampling, where Muon achieves lower generative perplexity at the same entropy level.
These results highlight the importance of optimizer choice. Nevertheless, models trained with both optimizers still outperform other baselines, suggesting that the strong performance of ELF cannot be attributed to the optimizer alone.

\subsection{Sampling Methods}
\label{sec:sampling_method}

\begin{figure}[t]
    \centering
    \includegraphics[width=0.75\linewidth]{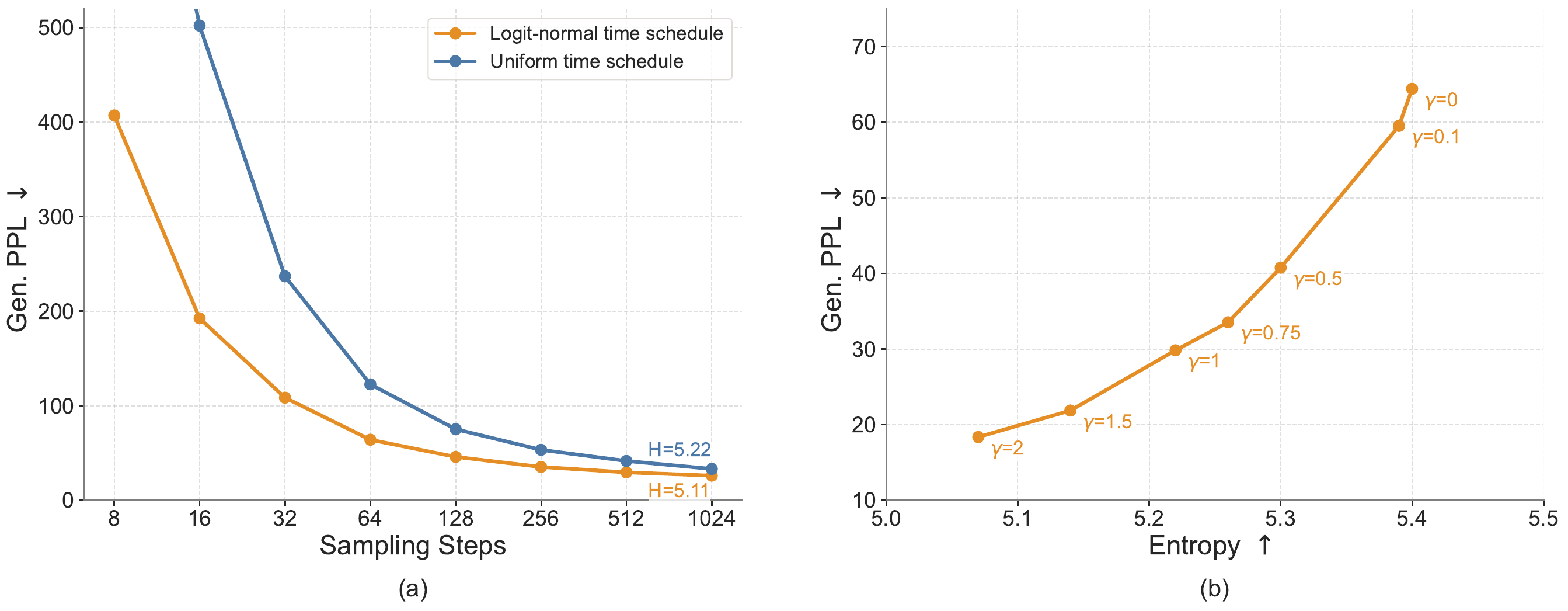}
    \caption{\textbf{Effect of time schedule and SDE noise re-injection scale.}
    (a) Logit-normal time schedule consistently improves generative perplexity across different sampling budgets, especially in the few-step regime.
    (b) The SDE noise re-injection scale $\gamma$ controls the generative perplexity--entropy trade-off by adjusting the amount of stochastic noise injected during sampling. 
    }
\label{fig:time_warping_and_sde}
\end{figure}

\begin{figure}[t]
    \centering
    \includegraphics[width=0.75\linewidth]{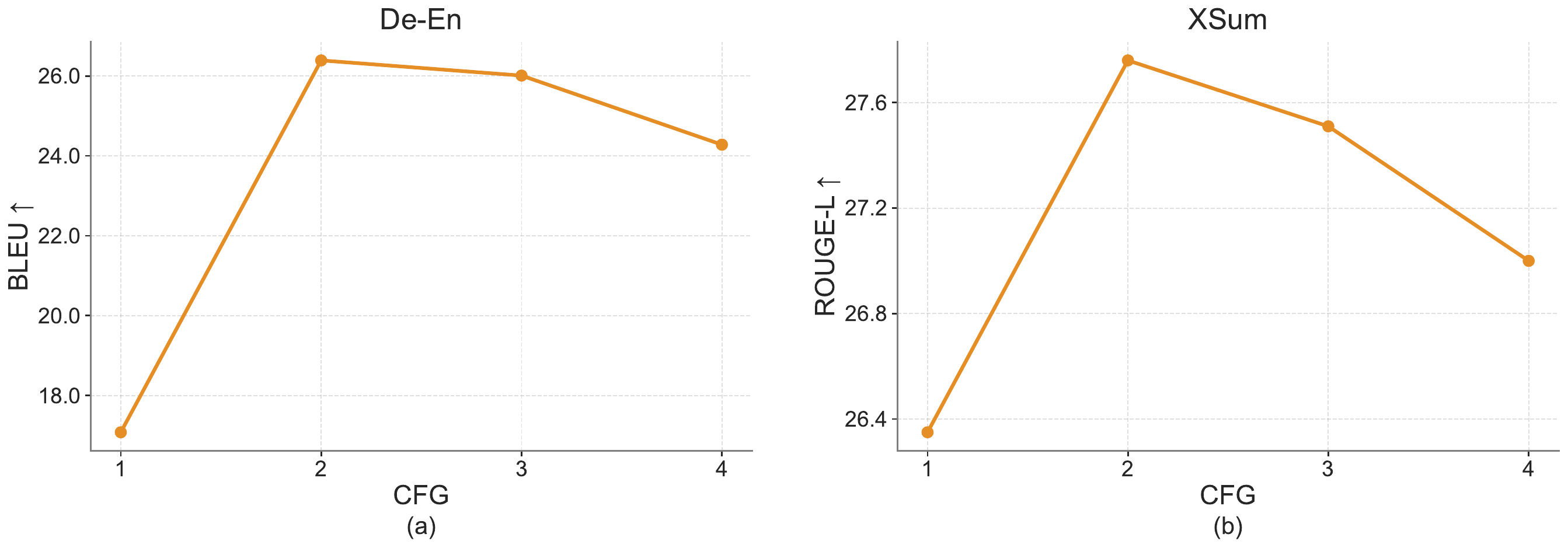}
    \caption{\textbf{Effect of CFG scale on conditional generation.}
    We sweep the CFG scale on WMT14 De-En translation and XSum summarization. Moderate guidance substantially improves task performance, with CFG scale 2 achieving the best result on both tasks, while overly strong guidance slightly degrades performance.}
\label{fig:cfg_conditional}
\end{figure}

We study two sampling design choices that improve inference efficiency and generation quality: sampling time schedule and stochastic SDE-inspired sampling. The logit-normal time schedule improves sampling efficiency by reducing the required number of denoising steps, while the SDE noise re-injection scale provides additional control over the generative perplexity--entropy trade-off.

\vspace{-0.5em}
\paragraph{Time schedules.}
By default, we use a logit-normal time schedule during inference~\citep{karras2022elucidating}. We also evaluate an alternative uniform schedule. Fig.~\ref{fig:time_warping_and_sde}a shows the effect of the time schedule on ODE sampling across different numbers of sampling steps. Across all step counts, the logit-normal schedule consistently reduces generative perplexity compared with the uniform schedule. This improvement is especially significant in the few-step regime. These results suggest that the logit-normal time schedule improves sampling efficiency and final sample quality, likely because it better aligns the inference-time trajectory with the training-time schedule and allocates more sampling steps to noisier time steps.

\vspace{-0.5em}
\paragraph{SDE noise re-injection scale.}
For SDE sampling, we introduce a noise re-injection scale hyperparameter $\gamma$ that controls the amount of stochasticity injected at each sampling step, as discussed in Sec.~\ref{sec:app_inference}. Intuitively, increasing $\gamma$ introduces more stochasticity, while $\gamma=0$ reduces to deterministic ODE sampling. As shown in Fig.~\ref{fig:time_warping_and_sde}b, $\gamma$ controls the generative perplexity--entropy trade-off: within a moderate range, larger $\gamma$ leads to lower generative perplexity while slightly reducing entropy. We hypothesize that the noise re-injection process helps correct early denoising errors, rather than deterministically amplifying imperfect trajectories as in ODE sampling. We therefore choose $\gamma=1.0$ as our default setting, which provides a strong balance between generative perplexity and entropy.

\subsection{CFG on Conditional Generation}
\label{sec:conditional_cfg}

We further study the effect of CFG scale on conditional generation tasks. As shown in Fig.~\ref{fig:cfg_conditional}, increasing the CFG scale from 1 to 2 substantially improves performance on both WMT14 De-En and XSum, suggesting that stronger conditioning helps the model better follow the source input. However, further increasing the scale leads to a gradual decline in performance, indicating that overly strong guidance can hurt generation quality. Based on this trend, we use CFG scale 2 as the default setting for conditional generation.
\FloatBarrier
\section{Experimental Details}
\subsection{Model Architecture}

Our model uses a standard Diffusion Transformer architecture~\citep{peebles2023scalable}. We also incorporate popular general-purpose improvements, including SwiGLU~\citep{shazeer2020glu}, RMSNorm~\citep{zhang2019root}, RoPE~\citep{su2024roformer}, and qk-norm~\citep{henry2020query}. We use in-context conditioning instead of adaLN-Zero~\citep{peebles2023scalable} conditioning, which allows us to significantly reduce the number of parameters; for example, the ELF-B model size is reduced from 148M to 105M parameters.
Tab.~\ref{tab:elf_scaling_hparams} summarizes the configurations of ELF across different model sizes. We report the Transformer depth, hidden size, number of attention heads, and parameter count. We also report the number of training epochs used on the OWT dataset for each variant. Larger models tend to learn faster in our setup, and therefore require fewer training epochs.

\begin{table}[t]
\centering
\caption{\textbf{ELF Model configurations} across different scales.}
\label{tab:elf_scaling_hparams}
\small
\setlength{\tabcolsep}{7pt}
\begin{tabular}{lccccc}
\toprule
\textbf{Model} & \textbf{Depth} & \textbf{Hidden size} & \textbf{\# Heads} & \textbf{Params} & \textbf{Training epochs} \\
\midrule
ELF-B & 12 & 768  & 12 & 105M & 5 \\
ELF-M & 24 & 1056 & 16 & 342M & 4 \\
ELF-L & 32 & 1280 & 16 & 652M & 3 \\
\bottomrule
\end{tabular}
\end{table}

\subsection{Hyperparameters}
\label{sec:app_hparams}

\begin{table*}[t!]
\centering
\caption{
\textbf{Default training hyperparameters} and setup for ELF-B on the OpenWebText dataset. Unless noted otherwise, all experiments in the paper follow this default configuration.}
\label{tab:training_hparams}
\small
\setlength{\tabcolsep}{6pt}
\renewcommand{\arraystretch}{1.12}
\begin{tabular}{@{}ll|ll@{}}
\toprule
\multicolumn{2}{c|}{\textbf{Model Architecture}} 
& \multicolumn{2}{c}{\textbf{Denoising and Decoding Config}} \\
\midrule
Model & ELF-B 
& Time schedule & logit normal \\

Model size & 105M 
& Denoiser $(P_{\text{mean}}, P_{\text{std}})$ & $(-1.5,\,0.8)$ \\

Encoder backbone & T5-small 
& Denoiser noise scale & $2.0$ \\

Embedding dimension & 512 
& Decoder $(P_{\text{mean}}, P_{\text{std}})$ & $(0.8,\,0.8)$ \\

Bottleneck dimension & 128 
& Decoder noise scale & $5.0$ \\

Model dimension & 768 
& Denoiser \vs decoder prob. & $0.8$ \vs $0.2$ \\

Sequence length & 1024 
& & \\

\midrule
\midrule
\multicolumn{2}{c|}{\textbf{Conditioning and Guidance}} 
& \multicolumn{2}{c}{\textbf{Optimization and Training}} \\
\midrule
Self-conditioning probability & $0.5$ 
& Optimizer & Muon \\

Self-conditioning CFG range & $[0.5,\,5]$ 
& Learning rate & $0.002$ \\

Num. of time tokens & 4 
& Weight decay & $0$ \\

Num. of model-mode tokens & 4 
& Training epochs & $5$ \\

Num. of CFG tokens & 4 
& Global batch size & 512 \\

SDE $\gamma$ & $1.0$ 
& Learning rate schedule & constant \\

& 
& Warmup epochs & $0.5$ \\

& 
& EMA decay & $0.9999$ \\

& 
& Training device & TPU v5p $\times$ 64 \\

& 
& Training time & 1.5 h per epoch \\
\bottomrule
\end{tabular}
\vspace{-1em}

\end{table*}
\paragraph{ELF pipeline hyperparameters.}
Tab.~\ref{tab:training_hparams} summarizes the main hyperparameters used in the ELF pipeline, covering model architecture, diffusion settings, conditioning and guidance, and optimization details. Unless noted otherwise, all experiments in the paper follow this default configuration. We include these settings for completeness and to facilitate reproducibility.

\vspace{-0.5em}
\paragraph{Inference-time settings for system-level comparison.}
For system-level comparison in Fig.~\ref{fig:system_level_comparison}, we use SDE sampling with time schedule enabled for all step budgets. We set the CFG scale to 3 for 8-, 16-, and 32-step generation. For SDE sampling, we use a stronger noise injection scale of $\gamma=2$ in the very few-step regimes of 8 and 16 steps, and reduce it to $\gamma=1.5$ for 32 steps, as longer denoising trajectories require less stochastic correction. For the system-level comparison in Tab.~\ref{tab:downstream}, we use 64-step ODE sampling with time schedule. We set the self-conditioning CFG scale to 1 and the input-condition CFG scale to 2.

\vspace{-0.5em}
\paragraph{Training-token budget for system-level comparison.}
Tab.~\ref{tab:training_tokens} reports the estimated effective training tokens used by ELF and each baseline in Fig.~\ref{fig:system_level_comparison}c. We estimate base-training tokens as $\text{batch size}\times\text{steps}\times\text{sequence length}$ and add distillation or flow-map stages on top where applicable. 
The OWT dataset contains roughly 9.04B tokens. With our default training schedule of 5 epochs, ELF therefore uses 45.2B effective training tokens. Thus, ELF requires roughly an order of magnitude fewer effective training tokens than the compared DLMs.

\begin{table}[t]
\centering
\caption{
\textbf{Estimated effective training tokens} for ELF and the prior DLM baselines used in our system-level comparison (Fig.~\ref{fig:system_level_comparison}c).
We estimate base-training tokens as $\text{batch size}\times\text{steps}\times\text{sequence length}$; distillation / flow-map stages are added on top where applicable.
}
\label{tab:training_tokens}
\small
\begin{tabular}{lllcc}
\toprule
\textbf{Method} & \textbf{Base training} & \textbf{Distillation  training} & \textbf{Effective tokens} & \textbf{Ratio} \\
\midrule
MDLM~\cite{sahoo2024simple}            & $512\times 1\text{M}\times 1024$ & -                                            & 524.3B & 11.6$\times$ \\
Duo~\cite{sahoo2025diffusion}          & $512\times 1\text{M}\times 1024$ & -                                            & 524.3B & 11.6$\times$ \\
MDLM\,+\,SDTT~\cite{sahoo2024simple,deschenaux2024sdtt}   & $512\times 1\text{M}\times 1024$ & $512\times 10\text{K}\times 5\times 1024$ & 550.5B & 12.2$\times$ \\
Duo\,+\,DCD~\cite{sahoo2025diffusion}  & $512\times 1\text{M}\times 1024$ & $512\times 10\text{K}\times 5\times 1024$ & 550.5B & 12.2$\times$ \\
FLM~\cite{lee2026one}                  & $512\times 1\text{M}\times 1024$ & -                                            & 524.3B & 11.6$\times$ \\
FMLM~\cite{lee2026one}                 & $512\times 1\text{M}\times 1024$ & $512\times 100\text{K}\times 1024$              & 576.7B & 12.8$\times$ \\
LangFlow~\cite{chen2026langflow}       & $512\times 1\text{M}\times 1024$ & -                                            & 524.3B & 11.6$\times$ \\
\midrule
\rowcolor[gray]{0.92}
\textbf{ELF (ours)}                    & $5\times 9.04\text{B}$ & -                                           & \textbf{45.2B} & \textbf{1.0$\times$} \\
\bottomrule
\end{tabular}
\end{table}

\subsection{Ablation Studies Setting}
\label{sec:ablation_studies_details}
We evaluate several choices of embedding representations for ELF and report the implementation details below. We also try two-stage training with a separate decoder.
Unless specified, we keep other settings the same as the default ELF configuration.

\vspace{-0.5em}
\paragraph{Scratch encoder.}
We train an encoder from scratch on OpenWebText~\citep{gokaslan2019OpenWeb} by following the original T5-small training pipeline~\citep{raffel2020exploring}. The encoder is trained for 5 epochs with a learning rate of $1\times10^{-3}$, cosine learning rate schedule, 0.4 epoch warmup, and a batch size of 512. During ELF training, we apply channel-wise normalization to the encoder outputs.

\vspace{-0.5em}
\paragraph{Pretrained embedding layer.}
We use the frozen embedding table from the T5-small encoder as the token embedding layer. The embedding layer matrix is normalized, and the unembedding layer is trained separately.

\vspace{-0.5em}
\paragraph{Gaussian embedding layer.}
We randomly initialize and freeze an embedding layer from a Gaussian distribution, with token-wise embedding mean 0 and standard deviation 1. The unembedding layer is trained separately using the decoder mode.

\vspace{-0.5em}
\paragraph{Learnable embedding layer.}
We jointly train the embedding layer together with the denoiser and decoder modes. The unembedding layer is tied with the embedding layer: denoiser-mode updates affect the embedding layer, while decoder-mode updates affect the unembedding layer. To stabilize training, we apply normalization directly on the unembedding layer matrix at every step.

\vspace{-0.5em}
\paragraph{Separate decoder.}
For the separate-decoder setting, we use a randomly initialized decoder architecture obtained by mirroring the T5-small encoder. 
We keep the encoder fixed, mask 20\% of the input tokens, and add logit-normal noise to the latent representations with $P_{\mathrm{mean}}=0.5$ and $P_{\mathrm{std}}=1.0$. 
The model is trained for 3 epochs with a learning rate of $3\times10^{-4}$ and a cosine learning-rate schedule. 
The relative noise scale with respect to the normalized latent representations is set to $5.0$.

\subsection{Reported Numbers}
\begin{table}[t]
\centering
\caption{\textbf{System-level ELF performance} reported as mean $\pm$ standard error (SE) over 6 independent evaluation runs (seeds 0--5; $n=6$).}
\label{tab:seed_summary_steps}
\small
\setlength{\tabcolsep}{8pt}
\begin{tabular}{rcccc}
\toprule
\textbf{Steps} & \textbf{SC CFG} &\textbf{$\gamma$} & \textbf{Gen. PPL $\downarrow$} & \textbf{Entropy $\uparrow$} \\
\midrule
8  & 3 & 2.0 & $67.32 \pm 2.25$ & $5.14 \pm 0.085$ \\
16 & 3 & 2.0 & $33.66 \pm 1.09$ & $5.16 \pm 0.026$ \\
32 & 3 & 1.5 & $24.08 \pm 0.16$ & $5.15 \pm 0.002$ \\
\bottomrule
\end{tabular}
\vspace{-2em}
\end{table}

\begin{table}[t]
\centering
\caption{\textbf{Scaling performance} of generative perplexity (Gen. PPL) and unigram entropy for ELF models of different sizes under SDE and ODE samplers with 64 sampling steps. The effect of self-conditioning (SC) CFG scaling diminishes beyond 3. 
}
\label{tab:elf_scaling_sampler_number}
\footnotesize
\setlength{\tabcolsep}{3.5pt}
\renewcommand{\arraystretch}{1.08}
\begin{tabular}{lccccccc}
\toprule
\multirow{2}{*}{\textbf{Sampler}} &
\multirow{2}{*}{\textbf{SC CFG}} &
\multicolumn{2}{c}{\textbf{ELF-B 105M}} &
\multicolumn{2}{c}{\textbf{ELF-M 342M}} &
\multicolumn{2}{c}{\textbf{ELF-L 652M}} \\
\cmidrule(lr){3-4}\cmidrule(lr){5-6}\cmidrule(lr){7-8}
& & \textbf{Gen. PPL} & \textbf{Entropy} & \textbf{Gen. PPL} & \textbf{Entropy} & \textbf{Gen. PPL} & \textbf{Entropy} \\
\midrule
\multirow{7}{*}{SDE}
& 0.5 & 36.77 & 5.28 & 39.21 & 5.35 & 37.50 & 5.41 \\
& 1.0 & 29.50 & 5.23 & 33.45 & 5.30 & 31.82 & 5.37 \\
& 1.5 & 25.25 & 5.18 & 28.42 & 5.26 & 28.72 & 5.35 \\
& 2.0 & 22.53 & 5.14 & 25.34 & 5.23 & 26.47 & 5.32 \\
& 3.0 & 19.72 & 5.10 & 21.69 & 5.18 & 23.31 & 5.28 \\
& 3.5 & \textcolor{gray!55}{37.56} & \textcolor{gray!55}{5.30} & \textcolor{gray!55}{36.48} & \textcolor{gray!55}{5.34} & 22.28 & 5.27 \\
& 4.0 & \textcolor{gray!55}{36.50} & \textcolor{gray!55}{5.29} & \textcolor{gray!55}{34.93} & \textcolor{gray!55}{5.33} & 21.37 & 5.26 \\
\midrule
\multirow{6}{*}{ODE}
& 0.5 & 104.29 & 5.51 & 88.51 & 5.51 & 68.27 & 5.52 \\
& 1.0 & 65.30 & 5.40 & 62.47 & 5.44 & 49.72 & 5.45 \\
& 1.5 & 44.85 & 5.31 & 46.71 & 5.37 & 39.97 & 5.40 \\
& 2.0 & 34.65 & 5.23 & 37.66 & 5.32 & 33.72 & 5.36 \\
& 3.0 & 26.62 & 5.15 & 28.80 & 5.24 & 26.57 & 5.29 \\
\bottomrule
\end{tabular}
\vspace{-1em}

\end{table}
\paragraph{System-level comparison.}
Across 6 independent evaluation seeds, ELF shows highly consistent system-level behavior, as shown in Tab.~\ref{tab:seed_summary_steps}. As the number of sampling steps increases from 8 to 32, the standard error (SE) decreases. The small standard errors—especially at 32 steps—suggest that these gains are robust to random seed variation and that the overall trend is reliable across runs. See Tab.~\ref{tab:seed_summary_steps} for detailed numbers.

\vspace{-0.5em}
\paragraph{Scaling behavior with CFG scales.}
The default setting for both sampling methods uses 64 sampling steps with the logit-normal time schedule. For the SDE sampler, we set $\gamma=1.0$. The exact numbers are reported in Tab.~\ref{tab:elf_scaling_sampler_number}. Larger CFG scales improve generation quality by reducing Gen. PPL within a certain range.
The effect of CFG scaling reverses beyond 3. Only ELF-L benefits from increasing the CFG scale from 3 to 4. Thus, in most default ablation studies, we only consider CFG scales from 0.5 to 3.

\begin{table}[b]
\centering
\caption{\textbf{Detailed training and evaluation configurations for conditional generation tasks} of our reproduced AR, MDLM, E2D2, and Duo baselines. AR, MDLM, and E2D2 are reproduced on XSum using the E2D2~\cite{arriola2025encoder} codebase with the Qwen3-0.6B tokenizer~\cite{yang2025qwen3} and follow the configurations reported in the E2D2 paper. For Duo, we build on the original Duo~\cite{sahoo2025diffusion} repository, add cross-attention conditioning and CFG, adapt the T5-small encoder to match our setting, and tune the hyperparameters to obtain the strongest reproduced results.}
\label{tab:all_configs_full}
\footnotesize
\setlength{\tabcolsep}{3.5pt}
\renewcommand{\arraystretch}{1.08}
\begin{tabular}{lccccc}
\toprule
\textbf{Config} & \textbf{AR} & \textbf{MDLM} & \textbf{E2D2} & \multicolumn{2}{c}{\textbf{Duo}} \\
\midrule
\multicolumn{6}{l}{\textit{Architecture}} \\
\midrule
Codebase                & E2D2 & E2D2 & E2D2 & Duo & Duo \\
Tokenizer               & Qwen3-0.6B & Qwen3-0.6B & Qwen3-0.6B & T5-small & T5-small \\
Hidden Size             & 256 & 256 & 256 & 768 & 768 \\
Intermediate Size       & 768 & 768 & 768 & -- & -- \\
\#Layers / Blocks       & 28 & 28 & enc=20, dec=8 & 12 & 12 \\
Sequence Length         & 64 & 64 & 64 & 64 & 64 \\
Max Cond Length         & 1024 & 1024 & 1024 & 1024 & 64 \\
Cond Embed         & -- & -- & -- & T5-small & T5-small \\
\midrule
\multicolumn{6}{l}{\textit{Training}} \\
\midrule
Dataset                 & XSum & XSum & XSum & XSum & De-En \\
Learning Rate           & 3e-4 & 3e-4 & 3e-4 & 3e-4 & 3e-4 \\
LR Scheduler            & const & const & const & const & const \\
Warmup  Steps                 & 1000 & 1000 & 1000 & 2500 & 2500 \\
Global Batch Size       & 128 & 128 & 128 & 512 & 512 \\
Optimizer               & DecoupledAdamW & DecoupledAdamW & DecoupledAdamW & AdamW & AdamW \\
Loss Type               & NLL & MDLM ELBO & E2D2 ELBO & Duo ELBO & Duo ELBO \\
Train Steps   & 500K & 500K & 500K & 1M & 1M \\
\midrule
\multicolumn{6}{l}{\textit{Evaluation}} \\
\midrule
Sampling Strategy         & greedy & predict\_and\_noise & predict\_and\_noise & Duo sampler & Duo sampler \\
Sampling Steps           & $L=64$ (AR) & $\approx L$ (first-hit) & $\approx L$ (first-hit) & 1000 & 1000 \\
Block size & 1 & 32 & 8 & - & - \\
CFG Scale                 & -- & -- & -- & 1.0 & 1.5 \\
Checkpoint                & best & best & best & best & best\\
EMA                       & true & true & true & true & true \\
\bottomrule
\end{tabular}
\vspace{-1em}
\end{table}

\subsection{Conditional Generation}
The WMT14 results for AR, MDLM, and E2D2 are taken from the E2D2~\cite{arriola2025encoder} paper, the DiffuSeq result is taken from the LD4LG~\cite{lovelace2023latent} paper, and the CDCD result is taken from the original CDCD~\cite{dieleman2022continuous} paper. For reproduced results, Duo~\cite{sahoo2025diffusion} is implemented using the Duo codebase\footnote{\url{https://github.com/s-sahoo/duo}}, while AR, MDLM, and E2D2 are reproduced using the E2D2 codebase\footnote{\url{https://github.com/kuleshov-group/e2d2}}.

We reproduce the baselines using settings adapted from their original implementations, as summarized in Tab.~\ref{tab:all_configs_full}. For AR, MDLM, and E2D2, we use the E2D2 codebase and adapt the CNN/DailyMail summarization configuration reported in the E2D2 paper to XSum. Note that although E2D2 is primarily designed for semi-autoregressive generation, we find that MDLM also achieves its best performance under a semi-autoregressive setting (\ie, block size 32 with two-block generation); using single-block diffusion without semi-autoregressive generation degrades performance. For Duo, we start from the official Duo repository and adapt it to our conditional generation setting by adding cross-attention conditioning and classifier-free guidance, and by using a T5-small encoder for the conditioning input. During inference, we generate without semi-autoregressive decoding. We tune the main sampling and guidance hyperparameters and report the best reproduced results we obtain.
\FloatBarrier
\lstdefinelanguage{none}{}
\newtcblisting{tcbverbatim}[1][]{
  blank, borderline={1pt}{-2pt},
  listing only,
  breakable,
  enhanced jigsaw,
  colback=white,
  left=10pt, right=10pt,
  toptitle=5pt,
  title={#1},
  fonttitle=\small\bfseries,
  coltitle=black,
  colbacktitle=gray!10,
  listing options={
    language=none,
    escapeinside={(*@}{@*)},
    basicstyle=\ttfamily\small,
    upquote=true,
    columns=flexible,
    breaklines=true,
    breakatwhitespace=true,
    breakautoindent=false,
    breakindent=0pt,
      literate=
        {’}{{'}}1
        {“}{{"}}1
        {”}{{"}}1
        {—}{{--}}1
        {£}{{GBP}}1
        {Â£}{{GBP}}1
        {ä}{{\"a}}1
        {ö}{{\"o}}1
        {ü}{{\"u}}1
        {Ä}{{\"A}}1
        {Ö}{{\"O}}1
        {Ü}{{\"U}}1
        {ß}{{\ss}}1
  }
}
\section{Qualitative Examples}
\label{sec:qualitative_examples}

\begin{figure}[t!]
    \centering
    \includegraphics[width=1.0\linewidth]{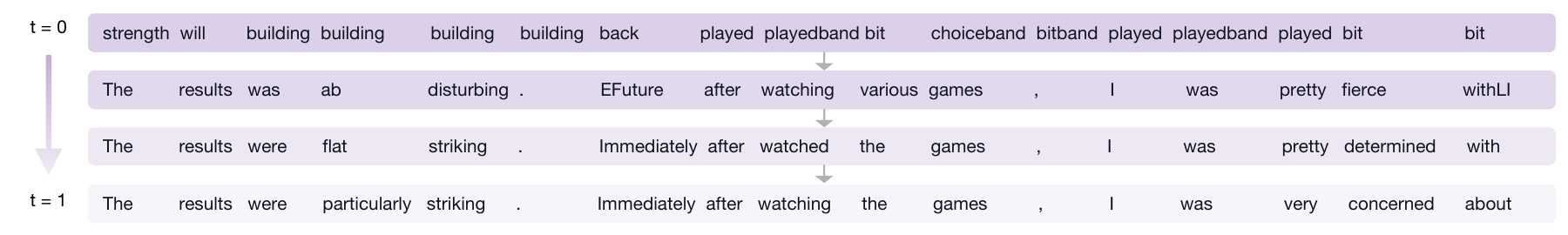}
    \caption{
\textbf{Denoising trajectory} of ELF-B. As $t$ increases from 0 to 1, ungrammatical sentences are progressively refined into fluent and grammatical text.
}
    \label{fig:denoising_trajectory}
\end{figure}
\subsection{Denoising Trajectory}
Fig.~\ref{fig:denoising_trajectory} visualizes the intermediate predictions along ELF's denoising process. Starting from repetitive tokens at $t=0$, the model gradually forms semantically meaningful phrases, improves grammar, and refines word choices as $t$ approaches 1. This trajectory illustrates how continuous diffusion generation progressively transforms noisy embeddings that decode to gibberish text into clean embeddings that decode to grammatical sentences.

\subsection{Unconditional Generation Examples on OpenWebText}

We provide three unconditional samples generated by ELF-B on OpenWebText, reported with their entropy and generative perplexity (Gen. PPL). The examples illustrate that ELF produces fluent, syntactically coherent, and topically consistent long-form text across diverse domains.

\vspace{0.5em}

\label{sec:qualitative_examples_uncon}
\begin{tcbverbatim}[ELF-B OWT]
Entropy: 5.36 Gen. PPL: 21.04

The company has been developing a virtual sleep mode for its iPhone and iPad for years. This means that users can improve their quality of life without turningping off their fingers thanks to Google's new virtual sleep technology. To make the experience a reality, virtual sleep mode was developed for Google, using a new built-in technology that includes real-time photography and shadow monitoring. This technology enables users to have a safe, comfortable look at where they sleep, even if you place the keyboard or a button under your fingers. Some sources point to the iPhone 6 and iPhone 6 as yet another example of the importance of virtual sleep mode in our everyday lives. This technology has been shown to be useful when staying busy on tight days, during difficult times or lying asleep on a hot night. This technology could also be used to improve sleep quality and help users improve quality of life. Editor's note: This post has been updated to answer to relevant questions. Google says it will add virtual sleep mode to its iPhone and iPad in coming week. Google announced some good news Thursday morning, including instructions on when to eat, checking out, where to sleep and the rules surrounding what to eat. The company reported a revenue of \$957 billion -- more than a third of the total revenue during the same period. But the company doesn't seem to have a slew of other good news yet, like the first one ...
\end{tcbverbatim}

\vspace{.5em}
\begin{tcbverbatim}[ELF-B OWT]
Entropy: 5.27 Gen. PPL: 21.29

Balin said the potential cost of starting there is very low, and he told USA2 Network in an interview that he is not only interested in expanding the capacity of the university, but is also interested in expanding other services, including student assistance, community assistance, youth assistance, youth assistance, and social justice assistance. Balin said, \"One of the things about this is that it's difficult to start, because if you're underfunded, you're going to need all the services that you need, and that's what you have to pay for. And it's going to difficult, if nothing, for you to get the funding you need to start right there.\" The UDU has not made such promise. \"We have a lot of the guys in the department that are doing well, a lot of the guys that aren't doing well at the university and they're currently underfunded,\" Balin said. \"Most of the other LHS universities across the country are currently underfunded. So, what do you want them to do? You know, right now, the cost is very low and there are no great universities in the rest of the country. It's not going to be easy.\" In the meantime, Balin said, the UDU isn't looking to attract high-quality...
\end{tcbverbatim}
\vspace{.5em}
\begin{tcbverbatim}[ELF-B OWT]
Entropy: 5.17 Gen. PPL: 21.80

Hey, I grew up in Lyndon in the early ’90s and, after my father’s death, began writing a book about himself called The Life of Steve O’Malse. After his second year at the University of Chicago O’Malse decided to write a biobio about his father. He finished his study at the University of Chicago in the fall of 1999. In 2009 he published a biobio called “My Dad And Daughter While he Was Home,” a successful biobio written by a former military officer, Lt. Gen. David Wilde. Steve O’Malse has had great national security experience. Throughout his career as a high-level national security adviser, he has served as an adviser to George H.W. Bush and an adviser to two top FBI officials, John J. Tillerson and Michael E. Comey, both of whom have been involved in the investigations that led to the resignation of Attorney General Rex Tillerson. He played a key role throughout the administration as a national security adviser, then as a special adviser to former President George W. Bush, then as secretary of Homeland Security under Bush and former presidential candidate Ronald Clinton. In 2008, O’Malse was named by President Ronald Reagan as a new national security adviser. In a speech last year, he detailed his experience in the Reagan administration, as a new national security adviser. O’Malse said he was surprised by his ability to express his concerns about national security, but added that he would be speaking more for years to come...
\end{tcbverbatim}

\subsection{Conditional Generation Examples}
\label{sec:qualitative_examples_con}

\paragraph{WMT14 De-En qualitative examples.}
We show qualitative examples on WMT14 De-En to complement the corpus-level BLEU results. ELF generally produces fluent and globally coherent translations.

\vspace{0.5em}

\begin{tcbverbatim}[ELF-B WMT-DE-EN]
<Original text>

Dieses Phänomen hat nach den Wahlen vom November 2010 an Bedeutung gewonnen, bei denen 675 neue republikanische Vertreter in 26 Staaten verzeichnet werden konnten.

<Reference translation>

This phenomenon gained momentum following the November 2010 elections, which saw 675 new Republican representatives added in 26 States.

<Our translation>

This phenomenon has increased in significance after the elections in November 2010, in which 675 new Republicanan representatives have been recorded in 26 countries.
\end{tcbverbatim}
\vspace{.5em}
\begin{tcbverbatim}[ELF-B WMT-DE-EN]
<Original text>

Im Gegensatz zu Kanada sind die US-Bundesstaaten für die Durchführung der Wahlen in den einzelnen Staaten verantwortlich.

<Reference translation>

Unlike in Canada, the American States are responsible for the organisation of federal elections in the United States.

<Our translation>

Unlike Canada, the United States are states responsible for holding elections in each country.
\end{tcbverbatim}

\paragraph{XSum qualitative examples.}
We show qualitative examples on XSum to complement the ROUGE results. ELF generally produces fluent and concise summaries that capture the main content of the source document.
\vspace{0.5em}

\begin{tcbverbatim}[ELF-B XSum]
<Original text>

Voges was forced to retire hurt on 86 after suffering the injury while batting during the County Championship draw with Somerset on 4 June.
Middlesex hope to have the Australian back for their T20 Blast game against Hampshire at Lord's on 3 August.
The 37-year-old has scored 230 runs in four first-class games this season at an average of 57.50.
"Losing Adam is naturally a blow as he contributes significantly to everything we do," director of cricket Angus Fraser said.
"His absence, however, does give opportunities to other players who are desperate to play in the first XI.
"In the past we have coped well without an overseas player and I expect us to do so now."
Defending county champions Middlesex are sixth in the Division One table, having drawn all four of their matches this season.
Voges retired from international cricket in February with a Test batting average of 61.87 from 31 innings, second only to Australian great Sir Donald Bradman's career average of 99.94 from 52 Tests.

<Reference summarization>

Middlesex batsman Adam Voges will be out until August after suffering a torn calf muscle in his right leg.

<Our summarization>

Middlesex captain Adam Voges will not miss the rest of the season as he struggles with a bone injury.
\end{tcbverbatim}
\vspace{.5em}
\begin{tcbverbatim}[ELF-B XSum]
<Original text>

Ms Kendall told the BBC Labour risked sending a "resignation letter to the British people as a serious party of government" by electing Mr Corbyn.
Separately, Ms Cooper warned there was a "serious risk the party will split" if the left-winger becomes its leader.
It comes as Labour begins sending out the first ballot papers to voters.
The result of the contest will be announced at a special conference on 12 September.
More than 600,000 people have signed up to vote in the four-way contest but Labour has said applications are still being verified.
610,753
total electorate, though this may fall as party removes those not entitled to vote
Of which, full party members: 299,755
Affiliated to a trade union: 189,703
Registered to vote by paying Â£3: 121,295
Meanwhile voting in the election for the new Scottish Labour leader ended at midday.
Mr Corbyn is due to unveil a 10-point policy plan while in Glasgow later.
The popularity of the left-wing Islington North MP, who is promising "a new kind of politics", has sparked a row about the future direction of the Labour party.
Another leadership contender, Andy Burnham, told the BBC Mr Corbyn's policies "lack credibility".
"It's not possible to promise free university education, re-nationalising the utilities, without that coming at a great cost and if you can't explain how that is going to be paid for then I don't think we'll win back the trust of voters on the economy," he said.
BBC political correspondent Ross Hawkins said there had been "frustration" in rival camps who accused Mr Burnham of being reluctant to take on Mr Corbyn. This appeared to be his most direct attack yet, he added.
But in an interview with Jeremy Vine on BBC Radio 2, Mr Burnham declined to follow Ms Kendall and Ms Cooper and advise his supporters not to back Mr Corbyn with their second and third preferences.
He added: "People will say if they hear things like that, 'hang on, what do you believe?'"
In an interview with The Independent, Ms Kendall called for voters to mark Ms Cooper or Mr Burnham as second and third preferences, and avoid giving votes to Mr Corbyn.
"I have set out very clearly where I differ with all the candidates but our differences with Jeremy's kind of politics are far greater," said Ms Kendall.
Speaking on BBC Radio 4's Today programme she said she "can't pretend to be agnostic" about a victory for Mr Corbyn, saying of the voting process: "It is an alternative vote system and I want to urge party members to use all of their different preferences.
"I will be using my second and third preferences and I would urge others to do the same because I don't want to see our party go back to the politics of the '80s, just being a party of protest."
The Leicester West MP also said she did not see the party splitting, as it did in the 1980s when Labour members formed the Social Democratic Party.
However, Ms Cooper told BBC 2's Newsnight: "I think there is a serious risk that the party will split, will polarise and I cannot bear to see that happen because there is too much at stake."
Asked in an interview with grassroots Labour website Labourlist whether voters should use their votes to try to prevent Mr Corbyn winning, she said: "I think people should use all of their preferences.
"And I think the focus has to be how do we make sure we can win that election, and that's the most important thing - and I don't think Jeremy can do that."
Mr Corbyn has warned against "personal abuse" in the campaign, saying he wants to focus on policy.
His policy programme includes a commitment to "growth not austerity", nationalising the railways and energy sector, and a plan for nuclear disarmament.
In an essay for the Fabian Society he also suggested Labour's new increased following should be more involved in the party and proposed a review of membership fees to make the party more "inclusive".
Former Prime Minister Gordon Brown is expected to join the debate over the leadership contest with a speech on Sunday, called "power for a purpose - the future of the Labour Party".
Lance Price, former director of communications for Labour, told the BBC the contest had been an "unedifying mess" and had "done nothing to reengage the labour party with those millions of people who deserted it".
The Guardian newspaper has endorsed Ms Cooper for the leadership while the Daily Mirror has given its backing to Mr Burnham, although the paper urged him to "find a role" in his team for Mr Corbyn, who it says has "lit up the election campaign".
Labour leadership hopefuls Liz Kendall and Yvette Cooper have said their supporters should back anyone other than Jeremy Corbyn in the contest.

<Reference summarization>

Labour leadership hopefuls Liz Kendall and Yvette Cooper have said their supporters should back anyone other than Jeremy Corbyn in the contest.

<Our summarization>

Labour leadership contender Kendall Kendall has warned she does not avoid an threat of using Jeremy Corbyn with second and third preferences.
\end{tcbverbatim}

\paragraph{GSM8K qualitative examples.}
We show qualitative examples on GSM8K to complement the accuracy results. For each problem, ELF-B generates a Python program, which is executed to obtain the final answer. ELF generally produces well-structured programs that restate the problem, define the given quantities, and reach the answer through a sequence of intermediate steps. The last example shows a failure case: the program computes the break-even point (12 years) rather than the first year in which the tree earns money (13 years).
\vspace{0.5em}

\begin{tcbverbatim}[ELF-B GSM8K]
<Question>

Judy teaches 5 dance classes, every day, on the weekdays and 8 classes on Saturday. If each class has 15 students and she charges $15.00 per student, how much money does she make in 1 week?

<Reference answer>

7425

<Our solution>

def simple_math_problem() -> int:
    '''
    Judy teaches 5 dance classes, every day, on the weekdays and 8 classes on Saturday.
    If each class has 15 students and she charges $15.00 per student, how much money does she make in 1 week?
    '''
    weekday_classes = 5
    saturday_classes = 8
    students_per_class = 15
    charge_per_student = 15.00
    total_classes = (weekday_classes * 5) + saturday_classes
    total_students = total_classes * students_per_class
    total_income = total_students * charge_per_student
    result = total_income
    return result

<Execution output>

7425.0 (correct)
\end{tcbverbatim}
\vspace{.5em}
\begin{tcbverbatim}[ELF-B GSM8K]
<Question>

Uriah's book bag is getting too heavy for him. He needs to remove 15 pounds from it. His comic books weigh 1/4 pound each and his toys weigh 1/2 pound each. If he removes 30 comic books, how many toys does he need to remove?

<Reference answer>

15

<Our solution>

def simple_math_problem() -> int:
    '''
    Uriah's book bag is getting too heavy for him.
    He needs to remove 15 pounds from it.
    His comic books weigh 1/4 pound each and his toys weigh 1/2 pound each.
    If he removes 30 comic books, how many toys does he need to remove?
    '''
    comic_book_weight = 1/4
    toy_weight = 1/2
    total_weight = 15

    comic_books_removed = 30

    total_comic_weight = comic_book_weight * comic_books_removed

    remaining_weight = total_weight - total_comic_weight

    toys_to_remove = remaining_weight / toy_weight

    result = toys_to_remove
    return result

<Execution output>

15.0 (correct)
\end{tcbverbatim}
\vspace{.5em}
\begin{tcbverbatim}[ELF-B GSM8K]
<Question>

4 adults and 8 children are to share 8 packets of chocolate bars. Each packet contains 5 chocolate bars. If each adult gets 6 chocolate bars and the rest are to be shared equally among the children, how many will each child get?

<Reference answer>

2

<Our solution>

def simple_math_problem() -> int:
    '''
    4 adults and 8 children are to share 8 packets of chocolate bars.
    Each packet contains 5 chocolate bars.
    If each adult gets 6 chocolate bars and the rest are to be shared equally among the children, how many will each child get?
    '''
    adults = 4
    children = 8
    packets = 8
    chocolate_bars_per_packet = 5
    chocolate_bars_per_adult = 6
    total_chocolate_bars = packets * chocolate_bars_per_packet
    adult_chocolate_bars = adults * chocolate_bars_per_adult
    remaining_chocolate_bars = total_chocolate_bars - adult_chocolate_bars
    bars_per_child = remaining_chocolate_bars // children
    result = bars_per_child
    return result

<Execution output>

2 (correct)
\end{tcbverbatim}
\vspace{.5em}
\begin{tcbverbatim}[ELF-B GSM8K]
<Question>

Carlos is planting a lemon tree. The tree will cost $90 to plant. Each year it will grow 7 lemons, which he can sell for $1.5 each. It costs $3 a year to water and feed the tree. How many years will it take before he starts earning money on the lemon tree?

<Reference answer>

13

<Our solution>

def simple_math_problem() -> int:
    '''
    Carlos is planting a lemon tree.
    The tree will cost $90 to plant.
    Each year it will grow 7 lemons, which he can sell for $1.5 each.
    It costs $3 a year to water and feed the tree.
    How many years will it take before he starts earning money on the lemon tree?
    '''
    cost_to_plant = 90
    lemons_per_year = 7
    selling_price_per_lemon = 1.5
    water_and_feed_per_year = 3
    total_revenue_per_year = lemons_per_year * selling_price_per_lemon
    profit_per_year = total_revenue_per_year - water_and_feed_per_year
    years_needed = cost_to_plant / profit_per_year

    result = years_needed

    return result

<Execution output>

12.0 (incorrect)
\end{tcbverbatim}

\subsection{Unconditional Generation Examples of ELF+PD on OpenWebText}
\label{sec:qualitative_examples_distill}

We show one sample from the final-round one-step distilled ELF+PD model at each sampling step, reported with its entropy and generative perplexity (Gen.\ PPL). 
\begin{tcbverbatim}[ELF+PD OWT, sampling steps = 1]
Entropy: 5.28 Gen. PPL: 132.89
Naturally Bloomberg has introduced a new Black Black S on the Laptop, which which the widely seems it it sound. — Ahan Geet (@Just AG) Uhr)) The The Black S Sport also well, Samsung has that it will likely to include its Hot Roller set.. 
Bull and want to are on the first Smartphone for A Mr Bols has came from his Apple web company ago ago ago ago, told my website earlier Tuesday that I will be waiting to see a new speedset soon. And he know if Apple will consider add with the mod and the software that the refurbished mini- Sculble. In the addition to this tablet, it offers a lot more quickly. 
Obviously, the first latest version of iPhone's the Sure Heartn't Version, although it's not considered the boosted design level of the OS many havesupposedly launched in Fia but, for However, there is unclear, but not unclear whether the is wills up for how the game will as soon as Software, translate. We'll a some people on to various pictures about, and I'll have a spring next week and watch a review on Samsung's iPhone Stream soon...
\end{tcbverbatim}

\vspace{.5em}
\begin{tcbverbatim}[ELF+PD OWT, sampling steps = 2]
Entropy: 5.47 Gen. PPL: 64.76

“I think that we have always had the responsibility to be able to make our environment more accessible to any users. WipWe have been truly one of the best gaming companies in the world. We’s been quite large in the years. Over the years, the company known, Wi-Breth has given them with a ideas on how to develop their own software. Despite the custom design design Dearms was unable to prove the company they wanted to become a portable programmeer, but then it took a lot of time that was quite ambitious and a little bit tedious. A littleerson, Vice manager of programming management at WipBreth and was someone who was helped with the development of a program known as OpenPreadrder. “We wanted to create a better, clean,, efficient solution, and then we could do it, that wasns not exactly what we wanted to,” Wiet said. “Our goal was to create a robust free loopor that allows live programs easily to your Windows devices. Now, this is a huge step because it’s fast and high-a programable, and because you don’t have to create full programs to all of your Windows devices, you definitely don’t have to beable to create everything for your Windows 10 devices, which you can’t really do. This will be a big step to your your own laptops, but it a lot more cleaner and more fun for the whole environment.” Conclusion: The entire version of Wi-Breth wastracted much attention by the developers and were hoping to develop their. decision: By the first time when after re got back into the project, they devised very OpenCrunter Engine and I am happy that they decided in they, they were able to create this tool. It is an extremely innovative, and would bring every functionality in the future. As very simple, it doesn’t have great a lot tweak support, but there is not much that can be required. WieBreal has incredible portability for use every any Windows Windows 10 devices. It works well with a large looper function, and can can be used all around the world in the future. “I think the project will have be completed but it’s not really a little difficult, at this point we hope to get it done again. We’ll have a to more more details in the future...
\end{tcbverbatim}

\vspace{.5em}
\begin{tcbverbatim}[ELF+PD OWT, sampling steps = 4]
Entropy: 5.32 Gen. PPL: 30.76

"I'm confident that I'll be the best player on the team, and I don't know how if you'll get him back as as long as he can. "He's amazing. I've scored one of 15 games of the season, and he played some games and did pretty well. I just take a lot of time to get back. He's a great young winger, and he doesn't have a lot of time left with him. He's got a great team, so he's a really good strikeer and he's got a lot of experience on team. But's squad is very good. He's a really, good guy." Spritley said that Reilly's being on the field this Saturday. "I don't know how he will be vs. Blues in Saturday, but I don't know. I don't know that. I just don't think he'll do anything. That's possible, though. I'm confident that." The U.S. Food and Drug Drug Agency is warning that there could be about a half billion people' use of eadain drug. The drug, known as Common Deperdomidicent's Disease (CCD), has become a popular of more than 1 million people over the past 10 years and has a range conditions such as hepat disease, cancer, obesity, and traumatic heart disease. However, eadain drug is already under development in many countries over the past decade. And, in a trend that's new, researchers are confident that eaedain drug can potentially create danger to many organisms and other medical diseases, such as heart cancer. In a study estimate, about 50\% of people using eaddain drug comes from other parts of the human body. Investigators believe eadain drug is that could reduce danger to life, but is not expected to lead to death. "We're expecting an up to two billion billion people of use over the next 10 or years,"...
\end{tcbverbatim}

\vspace{.5em}
\begin{tcbverbatim}[ELF+PD OWT, sampling steps = 8]
Entropy: 5.24 Gen. PPL: 19.22

Now let us take an deeper look at the decline in basic education programs over the last couple of decades. During the 1990s, when the basic industry was weakening, low-income people were declining. They didn’t have a good chance to learn it. But by the 1990s, basic education was weaker—low-income people were very lacking of money. They didn’t have the resources to understand the fundamental education program, which is a lack of adequate understanding of the program. What we’ve seen is the fact that the program is a political problem. The poor understanding of the fundamental program is a political problem for the low-income people. The basic program isn’t a politically political problem. The problem is that many people that education isn’t a political problem. It is not a political policy problem. They’re far less aware of how the fundamental program is being conducted. As we’ve seen, low-income people don’t get access to basic education. They don’t have enough scientific expertise and have enough of political experience. I’m a professor of politics in Minnesota and I am a professor of political science. I also talk about the fact that the majority of low-income people (65\%) don’t have the adequate money to learn to understand the basic program—which, I know, hasn’t happened. We need to become more aware that basic education for low-income people is declining. This is really an important part of making sure we have more understanding what’s behind these declines. When it comes to basic education, you should become more aware that the program isn’t based on knowledge of what’s happening and how it’s happening, so you can move on step further and get a little bit more sleep....
\end{tcbverbatim}\FloatBarrier

\end{document}